\documentclass[runningheads]{llncs}

\usepackage{eccv}

\usepackage{eccvabbrv}

\usepackage{graphicx}
\usepackage{booktabs}

\usepackage[accsupp]{axessibility}  

\usepackage{mathtools}
\usepackage{bbm}
\usepackage{pifont}
\usepackage{wrapfig}
\usepackage{algorithm}
\usepackage{algpseudocode}
\usepackage{tabularx}
\usepackage{graphics}
\usepackage{tabularray}
\usepackage{adjustbox}
\usepackage{multirow}

\UseTblrLibrary{booktabs}
\newcolumntype{P}[1]{>{\centering\arraybackslash}p{#1}}

\newcommand{\bs}{\boldsymbol}
\newcommand{\xmark}{\ding{55}}%

\def\measurehat#1{%
   \setbox0=\vbox{$\hat{#1}\hfil\break$\null\par
      \setbox0=\lastbox\unskip\unpenalty\global\setbox1=\lastbox}%
   \setbox0=\hbox{\unhbox1 \unskip\unpenalty\unskip \global\setbox2=\lastbox}%
   \setbox0=\vbox{\unvbox2 \setbox0=\lastbox}%
}
\def\doublehat#1{%
   \measurehat{#1}\dimen0=\wd0 \measurehat{\kern0pt#1}%
   \raise.35ex\rlap{\kern\dimexpr\dimen0-\wd0$\hat{\phantom{#1}}$}{\hat#1}%
}

\newtheorem{manualpropositioninner}{Proposition}
\newenvironment{manualproposition}[1]{%
  \renewcommand\themanualpropositioninner{#1}%
  \manualpropositioninner
}{\endmanualpropositioninner}

\newtheorem{manualcorollaryinner}{Corollary}
\newenvironment{manualcorollary}[1]{%
  \renewcommand\themanualcorollaryinner{#1}%
  \manualcorollaryinner
}{\endmanualcorollaryinner}

\newcommand{\beginsupplement}{
    \renewcommand*{\thesection}{\Alph{section}}
    \setcounter{section}{0}
    \setcounter{page}{1}
 }

\usepackage{hyperref}

\usepackage{orcidlink}

\begin{document}

\title{Self-Guided Generation of Minority Samples Using Diffusion Models}


\author{Soobin Um\orcidlink{0000-0002-1133-0027} \and
Jong Chul Ye\orcidlink{0000-0001-9763-9609}}

\authorrunning{S. Um and J. C. Ye}

\institute{KAIST, Daejeon, Republic of Korea\\
\email{\{sum,jong.ye\}@kaist.ac.kr}}

\maketitle

\begin{abstract}
We present a novel approach for generating minority samples that live on low-density regions of a data manifold. Our framework is built upon diffusion models, leveraging the principle of guided sampling that incorporates an arbitrary energy-based guidance during inference time. The key defining feature of our sampler lies in its \emph{self-contained} nature, \ie, implementable solely with a pretrained model. This distinguishes our sampler from existing techniques that require expensive additional components (like external classifiers) for minority generation. Specifically, we first estimate the likelihood of features within an intermediate latent sample by evaluating a reconstruction loss w.r.t. its posterior mean. The generation then proceeds with the minimization of the estimated likelihood, thereby encouraging the emergence of minority features in the latent samples of subsequent timesteps. To further improve the performance of our sampler, we provide several time-scheduling techniques that properly manage the influence of guidance over inference steps. Experiments on benchmark real datasets demonstrate that our approach can greatly improve the capability of creating realistic low-likelihood minority instances over the existing techniques without the reliance on costly additional elements. Code is available at \url{https://github.com/soobin-um/sg-minority}.
\end{abstract}

\section{Introduction}
\label{sec:intro}
Contemporary large-scale datasets often exhibit long-tailed distributions, containing \emph{minority} samples that lie on low-density regions of the data manifold. The minority samples are less common and often possess unique characteristics rarely seen in the majority of the data. Generating these less probable data points are indispensable in a variety of applications like classification~\cite{qin2023class}, anomaly detection~\cite{du2022vos, du2023dream}, and medical diagnosis~\cite{um2023don} where augmenting additional instances of rare attributes could enhance the predictive capabilities of the focused tasks. Such augmentation is also significant in promoting fairness, aligning with social vulnerabilities often associated with minority instances~\cite{um2023don, roh2023dr}. Moreover, the unique features within these minority instances are of paramount importance in use-cases like creative AI applications~\cite{rombach2022high, han2022rarity}, where the ability to generate samples with exceptional creativity is crucial.

\begin{figure}[!t]
    \begin{subfigure}[h]{0.54\textwidth}
    \centering
    \includegraphics[width=1.0\columnwidth]{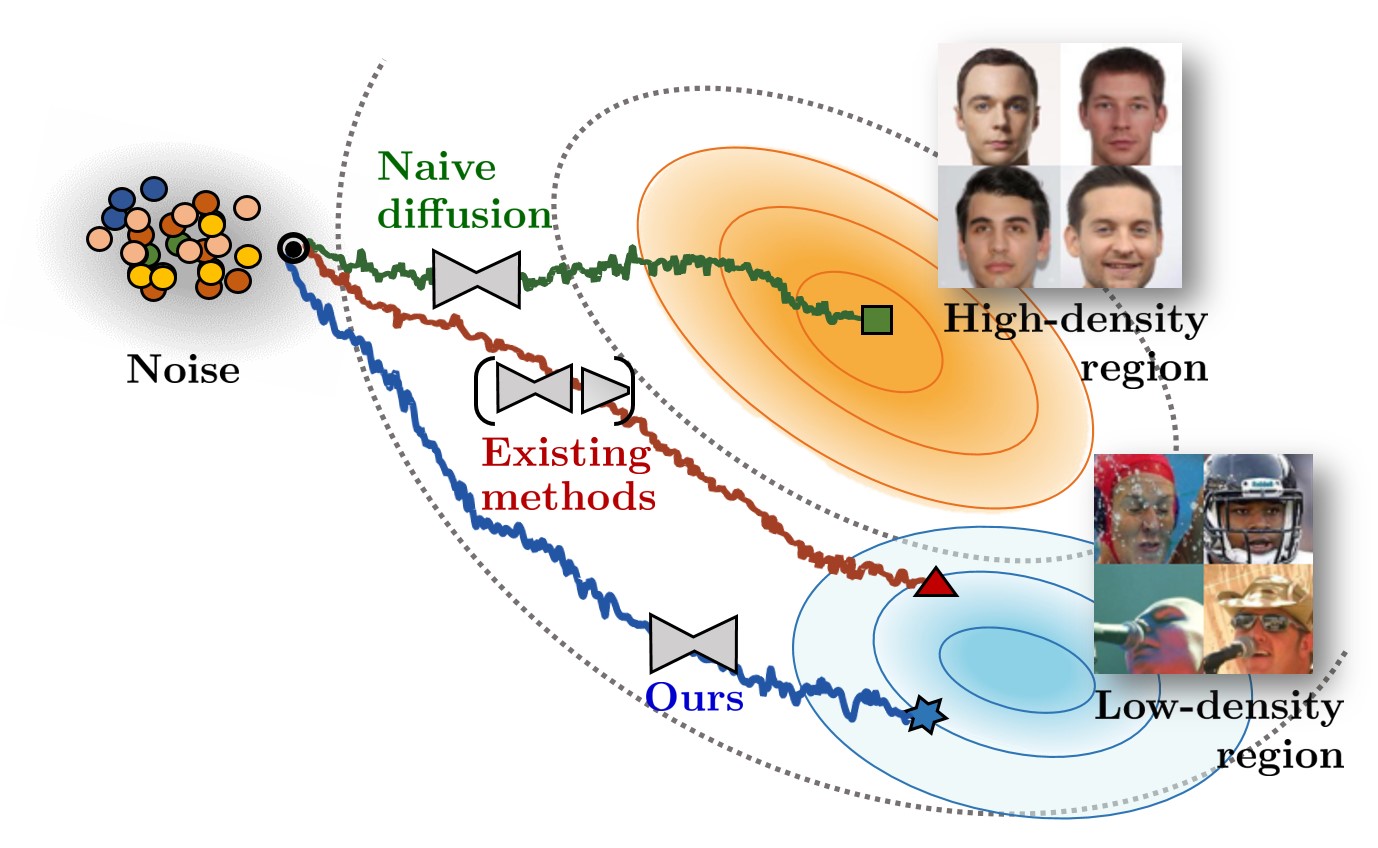}
    \end{subfigure}
    \hfill
    \begin{subfigure}[h]{0.44\textwidth}
    \centering
    \includegraphics[width=1.0\columnwidth]{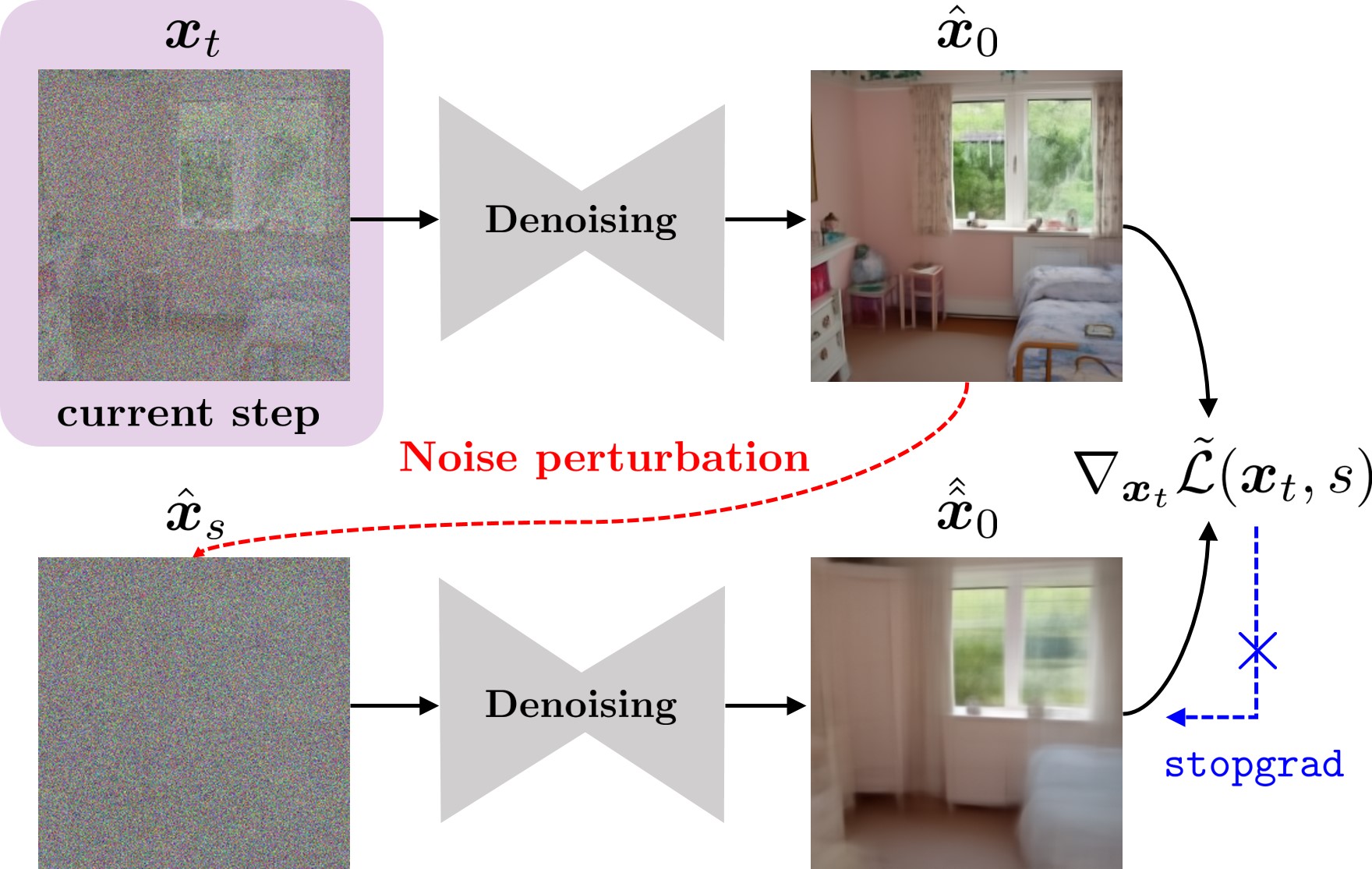}
    \end{subfigure}
\caption{\textbf{(Left) Existing methods vs. our self-guided approach.} Unlike previous methods that rely upon external components (\eg, classifiers) to guide the generation process towards low-density regions~\cite{sehwag2022generating, um2023don}, our approach yields low-density guidance solely based on a pretrained diffusion model, thereby offering a self-contained minority generation achievable without any aids of expensive extra elements. \textbf{(Right) Overview of our self guidance for minority data.} Specifically to yield low-density guidance given a current latent instance ${\bs x}_t$ during inference, we first obtain its denoised version $\hat{{\bs x}}_0$ via the use of Tweedie's formula~\cite{robbins1992empirical, chung2022diffusion} implemented with a pretrained model. We then perturb $\hat{{\bs x}}_0$ into $\hat{{\bs x}}_s$ via the DDPM forward process and denoise $\hat{{\bs x}}_s$ to $\doublehat{{\bs x}}_0$ via the pretrained model. A discrepancy between $\hat{{\bs x}}_0$ and $\doublehat{{\bs x}}_0$ (denoted as $\tilde{{\cal L}}({\bs x}_t, s)$ in the figure) is then computed, and we subsequently use its gradient as low-density guidance with the \texttt{stopgrad} technique~\cite{chen2021exploring} applied on $\doublehat{{\bs x}}_0$; see~\cref{sec:method} for details.}
\label{fig:overview}
\end{figure}

The problem is that under standard generative frameworks (\eg, GANs~\cite{goodfellow2014generative}, DDPM~\cite{ho2020denoising}), curating even a small number of low-probability samples requires substantial investments in terms of time and computational resources~\cite{hendrycks2021natural, sehwag2022generating, um2023don}. Several efforts were made to address this issue particularly under supervised settings~\cite{yu2020inclusive, lin2022raregan, sehwag2022generating, qin2023class}. However, most of their methods bear an inherent reliance on annotations (\eg, minority labels) that are difficult to obtain in many practically relevant scenarios~\cite{yu2020inclusive, lin2022raregan, sehwag2022generating, qin2023class}. Furthermore, they often require specialized training procedures~\cite{yu2020inclusive, lin2022raregan, qin2023class} which restrict the utilization of powerful pretrained models (like ADM~\cite{dhariwal2021diffusion} and Stable Diffusion~\cite{rombach2022high}), further limiting their practical significance in real-world applications.

One recent study by~\cite{um2023don} addressed the challenges using diffusion-based generative frameworks~\cite{sohl2015deep, ho2020denoising}. Their approach leverages a pretrained diffusion model and unlabeled data to implement a classifier-guided sampler that targets low-density regions by optimizing their own minority metric. While this method demonstrates substantial enhancements even under unsupervised settings, it often introduces significant overhead in obtaining the classifier particularly for large-scale benchmarks. For instance, training the classifier for the ImageNet-64 experiments in~\cite{um2023don} requires significant resource investments of \textbf{40} V100-days~\cite{dhariwal2021diffusion}. Also, the classifier construction necessitates access to a substantial number of real samples, which may be prohibitive in data-limited situations.

\noindent \textbf{Contribution.} In this work, we propose a \emph{minimalist} approach that eliminates the necessity for such expensive external components. Leveraging diffusion-based generative frameworks~\cite{sohl2015deep, ho2020denoising}, we develop a sampling technique that can be executed using only a pretrained model. One defining feature of our sampler lies in its \emph{self-guided} nature: unlike prior arts that obtain guidance toward low-density regions with the help of costly outside-components (\eg, separately trained classifiers~\cite{sehwag2022generating, um2023don}), our guidance for low-density regions is obtained in an \emph{autonomous} fashion by leveraging the knowledge of a given pretrained model; see~\cref{fig:overview} for an overview (the left plot). 

Specifically to identify the direction of a low-density region during inference time, we develop a metric that evaluates the uniqueness of features that are contained in an intermediate instance ${\bs x}_t$. More precisely, we employ a reconstruction loss w.r.t. the posterior mean $\mathbb{E}[ {\bs x}_0 | {\bs x}_t ]$, inspired by the low-likelihood measure proposed in~\cite{um2023don}. We provide both theoretical and empirical evidence that demonstrates a close connection between ours and existing likelihood measures for validation.

We then establish a sampler that optimizes the proposed metric over the generation process of diffusion models to encourage evolution towards low-likelihood minority features captured by our metric. We highlight that our method could be highly effective even with intermittent usage (\eg, incorporating our guidance every 5 sampling steps) thereby significantly reducing the computational costs associated with our approach. To improve the sample quality of generated minority samples, we further provide several scheduling methods that properly scale the strength of our guidance over the sampling timesteps.

We also conduct extensive experiments across a diverse range of real benchmarks. We demonstrate that the proposed sampler can significantly boost up the capability of generating high-quality minority samples, as reflected in high values of low-density metrics such as Average k-Nearest Neighbor and better quality metrics like Fr\'echet Inception Distance~\cite{heusel2017gans}. To highlight the practical significance of our work, we also delve into a downstream application, investigating the advantages of our sampler for data augmentation in classifier training. We emphasize that the benefits of our framework stem exclusively from a pretrained diffusion model, which is in stark contrast with existing techniques~\cite{yu2020inclusive, lin2022raregan, sehwag2022generating, qin2023class, um2023don} that require additional expensive resources to improve the minority-generating capability.

\noindent \textbf{Related work.} In addition to~\cite{um2023don}, the capability of producing minority data has been explored under several distinct conditions and scenarios~\cite{yu2020inclusive, lin2022raregan, sehwag2022generating, qin2023class, huang2023enhanced}. One close instance to our work is~\cite{sehwag2022generating} where the authors develop a diffusion sampler that can encourage the sampling process of diffusion models to evolve toward low-density regions w.r.t. a specific class via an external classifier and a class-conditional model. The key distinction w.r.t. our sampler is that their method requires access to the class predictor which is often expensive to obtain.

The authors in~\cite{yu2020inclusive, lin2022raregan, qin2023class, huang2023enhanced} investigate a slightly different goal, improving representations of minority instances to enhance data coverage close to the ground-truth data distribution. However, their methods rely upon label information that indicate minority samples, which is inherently distinct from our work.~\cite{samuel2023all} develops a sampler for text-to-image (T2I) diffusion models~\cite{rombach2022high} to specifically enhance the quality of generated samples prompted with unique concepts (\eg, shaking hands) rarely observed during training. However, their method requires access to a set of high-quality images that describe the target concepts and therefore not directly comparable to ours.

The exploration of diversity of diffusion models has received less attention compared to their quality aspects which have been scrutinized from various perspectives~\cite{ho2022classifier, hong2023improving}. One notable progress was recently made in~\cite{sadat2023cads} wherein the authors discovered that simply incorporating noise perturbations (yet properly annealed over time) to class embeddings could significantly improve the diversity of generated samples. A key difference from our approach lies in the fact that their method is designed to generate diverse samples that adhere to the ground-truth data distribution, rather than focusing on low-density regions of the distribution (as ours and~\cite{sehwag2022generating, um2023don}).


\section{Background}
\label{sec:background}


\subsection{Diffusion-based generative models}
\label{subsec:diff}

Diffusion models~\cite{sohl2015deep, ho2020denoising, song2019generative} are latent variable models described by a forward diffusion process and the associated reverse process. The forward process is basically a Markov chain with a Gaussian transition, where data is gradually perturbed by Gaussian noise according to a variance schedule $\{{\beta_t}\}_{t=1}^T$: $q({\bs x}_{t} | {\bs x}_{t-1}) \coloneqq {\cal N} ( {\bs x}_t ; \sqrt{1 - \beta_t} {\bs x}_{t-1}, \beta_t {\bs I})$ where $\{{\bs x}_t\}_{t=1}^T$ are latent variables with the same dimensionality as data ${\bs x}_0 \sim q({\bs x}_0)$. One important property of the forward process is that it admits \emph{one-shot} sampling of ${\bs x}_t$ at any timestep $t \in \{1, \dots, T\}$:
\begin{align}
\label{eq:DDPM_perturb}
q_{\alpha_t}({\bs x}_t \mid {\bs x}_0) = {\cal N} ({\bs x}_t; \sqrt{\alpha_t} {\bs x}_0, (1- \alpha_t) {\bs I}),
\end{align}
where $\alpha_t \coloneqq \prod_{s=1}^t (1 - \beta_s)$. The variance schedule is designed to respect $\alpha_T \approx 0$ so that ${\bs x}_T \sim {\cal N}(\bs{0}, \bs{I})$. The reverse process is another Markov Chain with learnable Gaussian transition $p_{\bs{\theta}}( {\bs x}_{t-1} | {\bs x}_t ) \coloneqq {\cal N}( {\bs x}_{t-1} ; \bs{\mu}_{\bs{\theta}}({\bs x}_t, t), {\bs \Sigma }_{\bs \theta}( {\bs x}_t, t )    )$. The mean is expressible in terms of a noise-conditioned score network as $\bs{\mu}_{\bs{\theta}}({\bs x}_t, t) = \frac{1}{\sqrt{1-\beta_t}}( {\bs x}_t + \beta_t {\bs s}_{\bs{\theta}}({\bs x}_t, t)  )$, where the score network is parameterized to approximate the score function of the perturbed distribution: ${\bs s}_{\bs{\theta}}({\bs x}_t, t) \coloneqq \nabla_{{\bs x}_t} \log p_{\bs{\theta}}({\bs x}_t)  \approx \nabla_{{\bs x}_t} \log q_{\alpha_t}({\bs x}_t) $. The variance of the reverse process is often fixed, \eg, ${\bs \Sigma}_{\bs \theta} ( {\bs x}_t, t )  = \beta_t {\bs I}$~\cite{ho2020denoising}. One common way to construct the score network is through a denoising score matching~\cite{vincent2011connection, song2020score}:
\begin{align*}
    \min_{{\bs \theta}} \sum_{t=1}^T w_t \mathbb{E}_{q({\bs x})q_{\alpha_t}(\tilde{{\bs x}} \mid {\bs x})}[ \| {\bs s}_{\bs{\theta}}(\tilde{{\bs x}}, t) 
    - \nabla_{\tilde{\bs x}} \log q_{\alpha_t}( \tilde{\bs x} \mid {\bs x}    )  \|_2^2     ],
\end{align*}
where $w_t \coloneqq 1 - \alpha_t$. One notable point is that this procedure is equivalent to training a noise-prediction network ${\bs \epsilon_{\theta}}({\bs x}_t, t)$ that predicts noise added on clean data ${\bs x}_0$ through the forward process in~\cref{eq:DDPM_perturb}~\cite{vincent2011connection, song2020score}. This establishes an intimate connection between the two networks: ${\bs s}_{\bs{\theta}}( {\bs x}_t, t )  =  -  {\bs \epsilon_{\theta}}({\bs x}_t, t) / \sqrt{1 - \alpha_t} $. Given a pretrained score model, the generation can be done by starting from ${\bs x}_T \sim {\cal N}({\bs 0}, {\bs I})$ and iteratively going through the reverse process down to ${\bs x}_0$:
\begin{align}
\label{eq:ancestral}
    {\bs x}_{t-1} =  \bs{\mu}_{\bs{\theta}}({\bs x}_t, t) + {\bs \Sigma }^{1/2}_{\bs \theta}( {\bs x}_t, t ) {\bs z}, \;\;\;{\bs z} \sim {\cal N}({\bs 0}, {\bs I}).
\end{align}
This process, often called \emph{ancestral sampling}~\cite{song2020score}, is actually a discretized simulation of a stochastic differential equation that defines $\{p_{\bs{\theta}}({\bs x}_t)\}_{t=0}^T$~\cite{song2020score}, which guarantees to sample from $p_{\bs{\theta}}({\bs x}_0) \approx q({\bs x}_0)$.


\subsection{Guided sampling with diffusion models}
\label{subsec:guided_sampling}

One instrumental feature of diffusion models is that their generative processes are often amenable to various optimization signals for conditioning generations in post-hoc fashions. Specifically at each time step $t$, one can incorporate an arbitrary energy-based \emph{guidance} into the sampling process (\eg,~\cref{eq:ancestral}) to encourage the evolution toward a desired direction~\cite{epstein2023diffusion}:
\begin{align}
\label{eq:guided}
    {\bs x}_{t-1} =  \bs{\mu}_{\bs{\theta}}({\bs x}_t, t) + {\bs \Sigma }^{1/2}_{\bs \theta}( {\bs x}_t, t ) {\bs z} + w_t {\bs g}({\bs x}_t, t),
\end{align}
where ${\bs g}({\bs x}_t, t)$ is a (sign-flipped) energy-based guidance function, and $w_t$ corresponds to the strength of the guidance term possibly scheduled over time. The guidance function may incorporate a target condition ${\bs c}$, in which case the function becomes ${\bs g}({\bs x}_t, t; {\bs c})$. Notice that plugging the gradient of a classifier log-likelihood (\eg. $\nabla_{{\bs x}_t}  \log p_{\bs \phi}(y | {\bs x}_t)$) into~\cref{eq:guided} (alongside $w_t = w {\bs \Sigma}_{\bs \theta} ( {\bs x}_t, t )$ where $w$ is a fixed constant) recovers the famous classifier-guided sampler~\cite{dhariwal2021diffusion}.

\noindent \textbf{Guidance for minority data.} The principles of existing minority samplers are centered around the classifier guidance~\cite{dhariwal2021diffusion}. Particularly for low-likelihood generation with a conditional diffusion model,~\cite{sehwag2022generating} propose to leverage the classifier guidance in the \emph{opposite} direction. Their guidance function is expressible as:
\begin{align*}
    {\bs g}({\bs x}_t, t; y) = -\nabla_{{\bs x}_t}  \log p_{\bs \phi}(y | {\bs x}_t).
\end{align*}
where $y$ indicates a target class for the focused conditional generation. The descending gradient makes the sampling process get closer to low-likelihood regions (w.r.t. the target class $y$), thereby encouraging generation of low-probability instances of the focused class $y$. On the other hand, the guidance developed by~\cite{um2023don} uses the same sign of the guidance as~\cite{dhariwal2021diffusion} while incorporating a distinct classifier, specifically trained to predict the degree of uniqueness of features within ${\bs x}_t$:
\begin{align}
\label{eq:mg}
    {\bs g}({\bs x}_t, t; l) = \nabla_{{\bs x}_t}  \log p_{\bs \psi}(l | {\bs x}_t),
\end{align}
where $l$ indicates the uniqueness level w.r.t. noisy latent instance ${\bs x}_t$. Their focused uniqueness metric, which is called \emph{minority score}, is shown as being inversely correlated with the likelihood (\ie, higher minority score, lower the likelihood)~\cite{um2023don}, and therefore the gradient ascent w.r.t. the metric can serve to encourage generation of highly unique (\ie, less-probable) instances.

While both techniques offer great improvements in the capability of producing minority instances~\cite{sehwag2022generating, um2023don}, their guidance functions bear inherent reliance on external components (like classifiers) that are often challenging (or even impossible) to acquire in practical settings. Our primary contribution lies in untethering such intrinsic dependencies and developing a \emph{self-contained} guidance function implementable solely through a pretrained diffusion model, thereby significantly enhancing the accessibility and practicality of minority instance generation.


\section{Method}
\label{sec:method}


\subsection{Towards an inference-time minority metric}
\label{subsec:M1}

Our approach starts by investigating a metric to be incorporated in the guidance function (\ie, ${\bs g}$ in~\cref{eq:guided}). Specifically in the context of self-guided generation of minority data, the metric should satisfy two criteria: (i) the ability to assess the likelihood of features underlain in an intermediate latent sample ${\bs x}_t$; (ii) accessibility via a pretrained diffusion model.

One can naturally think of leveraging an ODE-based likelihood estimator~\cite{song2020score} to compute $\log p_{\bs \theta} ({\bs x}_t)$ and incorporating the estimate in the guidance function. However, despite its capability of providing highly-accurate estimates, ODE-based estimators are often computationally expensive~\cite{song2020score}, \eg, requiring many Jacobian computations proportional to the number of diffusion timestep $T$. More importantly, the direct use of the log-likelihood in the guidance function (\eg, $g({\bs x_t}, t) = \nabla_{{\bs x}_t} \log p_{\bs \theta} ({\bs x}_t)$) may drive the sampling process \emph{out-of-manifold}. This is because a low-likelihood in a perturbed distribution may imply a noisy instance that does not belong to the data manifold. The downside is evident by poor performance of the high-temperatured sampler of diffusion models; see details in Sec. G in~\cite{dhariwal2021diffusion}.

We take a distinct approach that sidesteps the above challenges. To this end, we first introduce \emph{minority score}, a low-likelihood measure proposed in~\cite{um2023don}. The metric quantifies the degree of uniqueness (\ie, low-densitiness) of features contained in a given \emph{clean} sample ${\bs x}_0$, mathematically written as:
\begin{align}
\label{eq:ms}
    {\cal L}({\bs x}_0; t) \coloneqq 
    \mathbb{E}_{q_{\alpha_t} ({\bs x}_t \mid {\bs x}_0) } [ d( {\bs x}_0, \hat{{\bs x}}_0 ({\bs x}_t) ) ],
\end{align}
where $t$ refers to the timestep used for perturbing ${\bs x}_0$, and $d(\cdot, \cdot)$ is a discrepancy measure (\eg, LPIPS~\cite{zhang2018perceptual}). $\hat{{\bs x}}_0$ denotes the posterior mean obtained via Tweedie's formula~\cite{robbins1992empirical, chung2022diffusion} implemented with a pretrained model ${\bs s}_{\bs \theta}$:
\begin{align}
\label{eq:tweedie_with_score}
    \hat{{\bs x}}_0 ({\bs x}_t)  \coloneqq \mathbb{E}[{\bs x}_0 | {\bs x}_t] = \frac{1}{\sqrt{\alpha_t}}\left( {\bs x}_t + (1 - \alpha_t)   {\bs s}_{\bs \theta} ({\bs x}_t, t)   \right).
\end{align}
Intuitively, minority score can be interpreted as a reconstruction loss of a clean sample, measured with the posterior mean of a noise-perturbed version. The key benefit of this metric is in its computational efficiency, requiring only a single forward pass of a diffusion model while serving as a good proxy for the log-likelihood~\cite{um2023don}. The problem is that the metric is defined w.r.t. clean samples ${\bs x}_0$, making it impossible to be directly employed in the guidance function (that should work with ${\bs x}_t$). The authors in~\cite{um2023don} circumvented this issue by introducing a separately-trained classifier (\eg, in~\cref{eq:mg}), but as mentioned earlier, it is often expensive to obtain.

\begin{figure}[t]
    \centering
    \includegraphics[width=1.0\columnwidth]{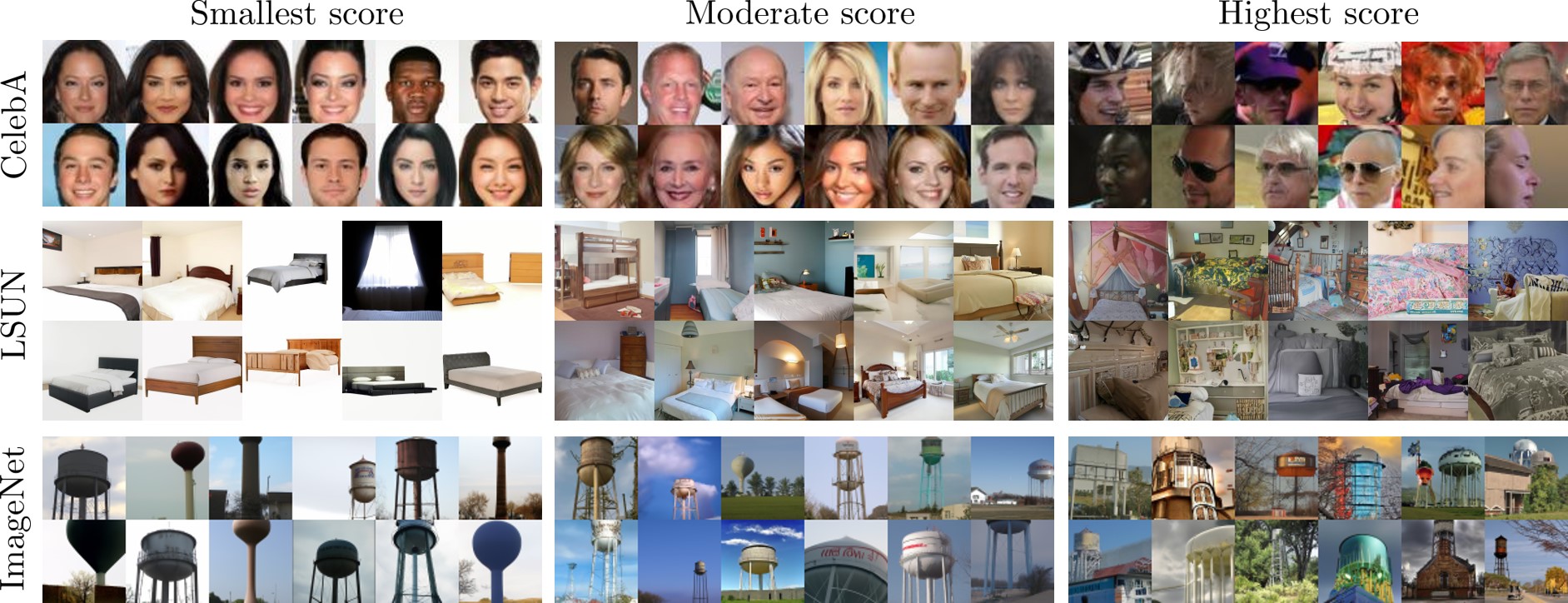}
    \caption{Effectiveness of our metric (\ie,~\cref{eq:mse}) for identifying low-likelihood minority features during inference time. Generated samples with the smallest (left column), moderate (middle column), and the highest (right column) values of the proposed metric are exhibited. We employed ADM~\cite{dhariwal2021diffusion} for the generations of all three benchmarks. The metric values were calculated during inference time via~\cref{eq:mse}.}
    \label{fig:mse_top_bottom}
\end{figure}

We make progress by providing a new metric that accommodates noisy intermediates ${\bs x}_t$. The key idea is to consider the posterior mean $\hat{\bs x}_0 $ as a clean-surrogate of ${\bs x}_t$ and to measure the uniqueness of features therein. Specifically we employ a reconstruction loss of $\hat{{\bs x}}_0$:
\begin{align}
\label{eq:mse}
    \tilde{{\cal L}} ( {\bs x}_t; s ) \coloneqq 
    \mathbb{E}_{q_{\alpha_s} ( \hat{{\bs x}}_s \mid \hat{{\bs x}}_0) }
    [ d(  \hat{{\bs x}}_0 ( {\bs x}_t ),  \doublehat{{\bs x}}_0 ( \hat{{\bs x}}_s ({\bs x}_t) )  ) ],
\end{align}
where $s$ refers to the timestep used for noise-corrupting $\hat{{\bs x}}_0$, and $\hat{\bs x}_s$ indicates the perturbed instance: $\hat{\bs x}_s( {\bs x}_t ) \coloneqq   \sqrt{\alpha_s}\hat{{\bs x}}_0 ( {\bs x}_t ) +  \sqrt{ 1 - \alpha_s } {\bs \epsilon}  $. $\doublehat{{\bs x}}_0 ( \hat{{\bs x}}_s )$ denotes the posterior mean of $\hat{{\bs x}}_s$ obtained by applying~\cref{eq:tweedie_with_score} on $\hat{{\bs x}}_s$. $d(\cdot, \cdot)$ is a distance measure that plays the same role as in~\cref{eq:ms}. See the supplementary for details on our choices of $s$ and $d(\cdot, \cdot)$. Note that our metric can be understood as minority score of $\hat{\bs x}_0({\bs x}_t)$, a variant that extends the metric to describe the uniqueness w.r.t. noisy instances ${\bs x}_t$.

\cref{fig:mse_top_bottom} visualizes the effectiveness of the proposed metric on several real-world benchmarks. Observe that the instances in the left column, which are determined as high-likelihood samples by our metric, exhibit features that are commonly observed in the corresponding datasets (\eg, frontal-view faces in CelebA~\cite{liu2015faceattributes}). While in the right column, we see samples containing uncommon visual attributes of the benchmarks. For instance, ``Wearing\_Hat'' and ``Eyeglasses'' attributes observed in the CelebA samples are famously known as minority features~\cite{amini2019uncovering, yu2020inclusive}. Also, complicated visual attributes seen in the LSUN and ImageNet samples are actually key defining features of low-density instances in real-world benchmarks~\cite{serra2019input, arvinte2023investigating}. In the supplementary, we illustrate the correlations of the proposed metric and existing low-density metrics; see details therein.

\noindent \textbf{Connection to log-likelihood.} To further validate our metric in a theoretical manner, we establish a mathematical connection of the proposed metric with log-likelihood. To this end, we first prove that minority score, when integrated over timesteps with a properly chosen $d(\cdot, \cdot)$, becomes equivalent to the negative Evidence Lower BOund (ELBO) -- a well-known proxy for log-likelihood. Then by leveraging the relation of our metric with minority score, we establish the connection with the ELBO. Below we provide formal statements of the claim. See the supplementary for the proofs.
\begin{proposition}
\label{proposition}
    Consider minority score in~\cref{eq:ms} with the squared-error distance loss $\| \cdot \|_2^2$. Its weighted sum over timesteps is equivalent (upto a constant factor) to the negative ELBO considered in~\cite{ho2020denoising}:
    \begin{align*}
        \sum_{t=1}^T \bar{w}_t {\cal L}({\bs x}_0; t)
         = \sum_{t=1}^T {\mathbb E}_{p({\bs \epsilon})}
        [ \| {\bs \epsilon} -  {\bs \epsilon}_{\bs \theta} ( \sqrt{\alpha_t} {\bs x}_0 + \sqrt{1 - \alpha_t} {\bs \epsilon}, t )  \|_2^2  ] \gtrapprox - \log p_{\bs \theta} ({\bs x}_0),
    \end{align*}
    where $\bar{w}_t \coloneqq \alpha_t / (1- \alpha_t) $ and $p ( {\bs \epsilon} ) \coloneqq {\cal N}({\bs \epsilon} ; {\bs 0}, {\bs I})$.
\end{proposition}

\begin{corollary}
\label{corollary}
    The proposed metric in~\cref{eq:mse} with the squared-error loss is equivalent to the negative ELBO w.r.t. $\log p_{\bs \theta} (\hat{{\bs x}}_0 ({\bs x}_t))$ when integrated over timesteps with $\bar{w}_s \coloneqq \alpha_s / (1- \alpha_s)$.
\end{corollary}


\subsection{Self-guidance for low-density regions}
\label{subsec:method_2}

Our next step is to develop the guidance function that incorporates our metric for minority generation. Since we are interested in encouraging ${\bs x_t}$ to evolve toward low-likelihood regions (that could yield high values of~\cref{eq:mse}), a natural choice for ${\bs g}$ would be to use the gradient of the proposed metric. Employing the gradient of our measure as ${\bs g}$ gives:
\begin{align}
\label{eq:our_g_wo_sg}
    {\bs g}({\bs x_t}, t; s) \coloneqq \nabla_{{\bs x}_t} \tilde{{\cal L}} ( {\bs x}_t; s ) = \nabla_{{\bs x}_t} \mathbb{E}_{q_{\alpha_s} ( \hat{{\bs x}}_s \mid \hat{{\bs x}}_0) } [ d(  \hat{{\bs x}}_0 ( {\bs x}_t ),  \doublehat{{\bs x}}_0 ( \hat{{\bs x}}_s ( {\bs x}_t ) )  ) ].
\end{align}
Notice that this guidance function does not require any external elements for computation, which is in stark contrast with the prior methods on low-density guidance~\cite{sehwag2022generating, um2023don}. We empirically found that simply adopting the above guidance function can yield great improvements in the capability to produce minority instances. However, the gradient computation in~\cref{eq:our_g_wo_sg} requires two backward passes through the model ${\bs s}_{\bs \theta}$, which often comes with considerable computational overhead.

We handle the issue by leveraging the stop-gradient technique~\cite{chen2021exploring}. More specifically, we employ the stop-gradient on $\doublehat{{\bs x}}_0$ that incurs the additional backward pass. Our modified guidance reflecting the stop-gradient reads:
\begin{align*}
    {\bs g}^*({\bs x_t}, t; s) \coloneqq \nabla_{{\bs x}_t} \mathbb{E}_{q_{\alpha_s} ( \hat{{\bs x}}_s \mid \hat{{\bs x}}_0) } [ d(  \hat{{\bs x}}_0 ( {\bs x}_t ),  \texttt{sg}(\doublehat{{\bs x}}_0 ( \hat{{\bs x}}_s ( {\bs x}_t ) )  ) ) ],
\end{align*}
where $\texttt{sg}(\cdot)$ indicates the stop-gradient operator. Notice that only a single backward pass now suffices for computing the gradient. Importantly, we found that it often preserves the great performance benefit of low-likelihood guidance offered by the guidance function in~\cref{eq:our_g_wo_sg}. Conversely, incorporating the stop-gradient on $\hat{\bs x}_0$ (instead of $\doublehat{{\bs x}}_0$) is not as effective, yielding little improvements over standard diffusion samplers. We conjecture that this trend is because the impact of concerning $\hat{\bs x}_0$ is much more significant in optimizing the proposed metric than the influence w.r.t. $\doublehat{{\bs x}}_0$ which is less relevant to the current sample ${\bs x}_t$ due to the perturbation ${\bs \epsilon}$. See~\cref{table:impact_sg} for empirical validation regarding our stop-gradient approach. An overview of our guidance can be found in~\cref{fig:overview} (the right-plot). We found that our guidance could be robust to the off-manifold issue (mentioned in~\cref{subsec:M1}); see the supplementary for a detailed analysis on this point.

\noindent \textbf{Intermittent guidance for reduced computations.} We discovered that the computational costs of our sampler could be significantly reduced through \emph{intermittent} usage, \ie, incorporating the guidance once every $n$ sampling steps. For instance, employing the guidance every 2 steps (\ie, $ n = 2 $) could lead to a notable reduction of 37.7\% in inference time with marginal impact on performance (see~\cref{table:metrics} for instance). Empirically, we observed that employing $ n = 5 $ yields satisfactory performance across diverse benchmarks. See the supplementary for detailed analysis on the impact of $n$.


\subsection{Time-scheduling for improved sample quality}
\label{subsec:method_3}

\begin{wrapfigure}{r}{0.55\textwidth}
\vspace{-4.2em}
\begin{minipage}{0.55\textwidth}
\begin{algorithm}[H]
    \caption{Self-guided minority sampler}
    \label{alg:sampling}
    \begin{algorithmic}[1]
        \Require{$T, n, s, w$.}
        \State{${\bs x}_T \sim {\cal N}({\bs 0}, {\bs I})$}
        \For{$t \gets T$ to $1$}
            \State{${\bs z} \sim {\cal N}({\bs 0}, {\bs I})$ if $t>1$, else ${\bs z} = {\bs 0}$}
            \State{${\bs x}_{t-1} \gets  \bs{\mu}_{\bs{\theta}}({\bs x}_t, t)  + {\bs \Sigma }^{1/2}_{\bs \theta}( {\bs x}_t, t ) {\bs z}$}
            \If{ $t \bmod n = 0$ }
            \State{${\bs x}_{t-1} \gets {\bs x}_{t-1} + w  {\bs \Sigma}_{\bs \theta} ( {\bs x}_t, t ) {\bs g}^*({\bs x_t}, t; s) $}
            \EndIf
        \EndFor
        \item[]
        \Return{${\bs x}_0$}
    \end{algorithmic}
\end{algorithm}
\end{minipage}
\vspace{-2.2em}
\end{wrapfigure}

Now we move onto the story on $w_t$, a scaling factor that controls the strength of the guidance over time. A naive approach is to use a constant scale (\ie, $w_t = w$), but we observed that it often leads to non-trivial degradation in sample quality for high values of $w$. We hypothesized that this comes from conflicting influences between the reverse process and our guidance, particularly occurring during the later timesteps. Specifically, the sampling process in these later steps often focuses on articulating fine details of images~\cite{kim2021soft, choi2022perception, karras2022elucidating}. If our guidance remains consistently strong during these stages, it may impede the articulation process since our guidance could potentially encourage structural changes diverging from the refinement task. To avoid the conflict, we explored several time-scheduled scaling methods that employ decreasing $w_t$'s over time. We found that they exhibit the same trend of yielding better sample quality over constant scales with a slight compromise in the low-densitiness.

\begin{figure}[!t]
\begin{subfigure}[h]{0.321\textwidth}
    \centering
    \includegraphics[width=1.0\columnwidth]{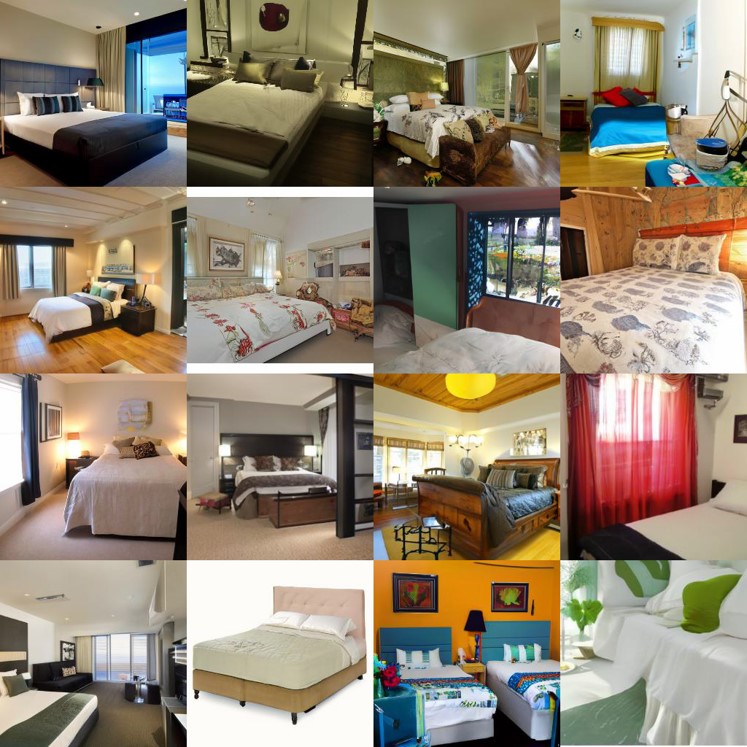}
    \caption{ADM~\cite{dhariwal2021diffusion}}
    \end{subfigure}
    \hfill
    \begin{subfigure}[h]{0.321\textwidth}
    \centering
    \includegraphics[width=1.0\columnwidth]{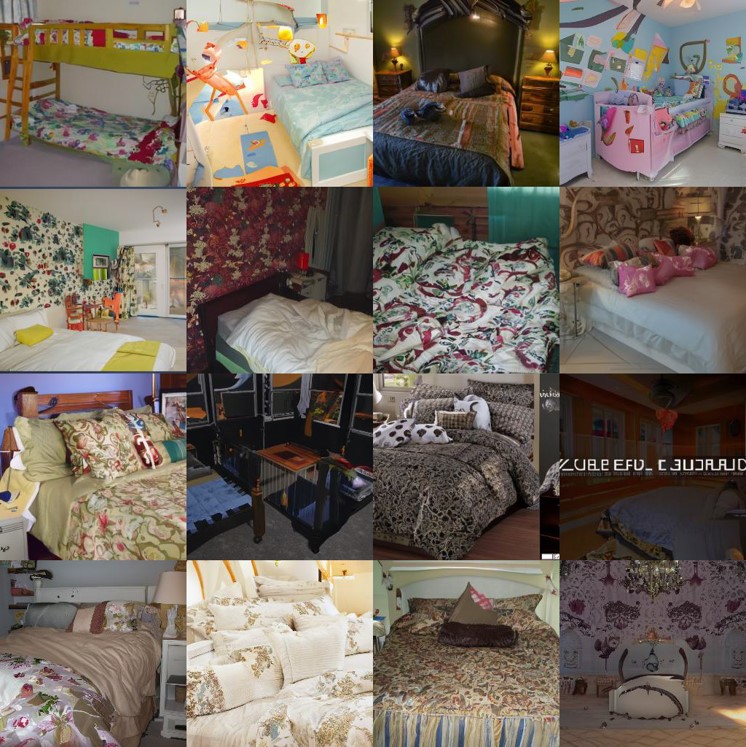}
    \caption{Um and Ye~\cite{um2023don}}
    \end{subfigure}
    \hfill
    \begin{subfigure}[h]{0.321\textwidth}
    \centering
    \includegraphics[width=1.0\columnwidth]{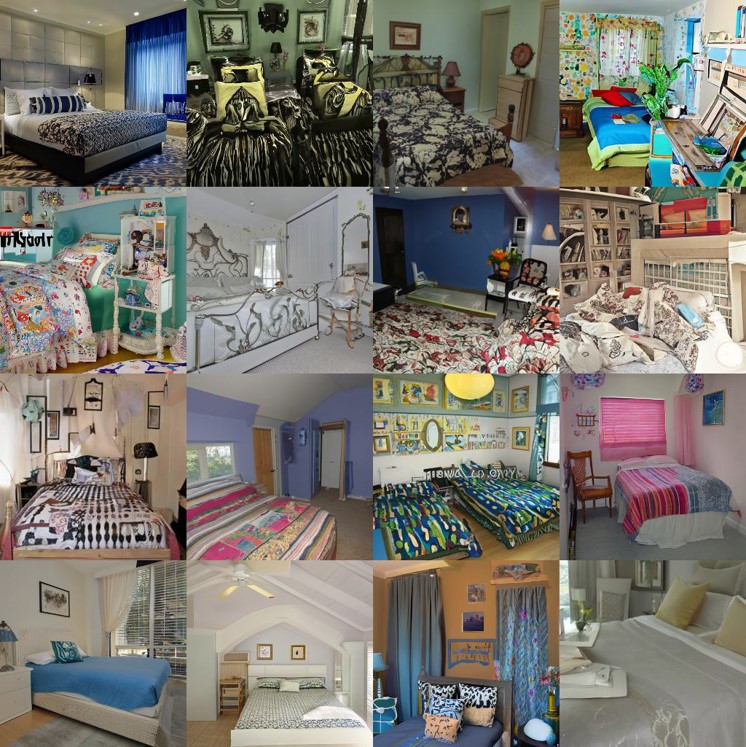}
    \caption{Ours}
\end{subfigure}
\caption{Sample comparison on LSUN-Bedrooms. We share the same random seed for all methods.}
\label{fig:lsun_samples}
\end{figure}

We provide two distinct time schedules herein. The first one is a simple switch-off-type schedule that discontinues incorporating the guidance after a specific timestep:
\begin{align*}
    w_t = w \cdot \mathbbm{1}\{t \ge t_{\text{mid}} \},
\end{align*}
where $t_{\text{mid}}$ is a pre-defined timestep that determines when to stop. The other proposal is one that leverages the noise variance of the reverse process (\ie, the same choice as~\cite{dhariwal2021diffusion}):
\begin{align*}
    w_t = w \cdot {\bs \Sigma}_{\bs \theta} ( {\bs x}_t, t ),
\end{align*}
where ${\bs \Sigma}_{\bs \theta} ( {\bs x}_t, t )$ can be either learned or fixed; see~\cref{subsec:diff} for its formal definition. While the switch-off yields some improvements over the sampler with fixed scales, we empirically observed that the variance-based schedule generally yields better performance; see~\cref{table:impact_w_t} for instance. The proposed minority sampler with the variance schedule is formulated as:
\begin{align}
\label{eq:guided_ours}
    {\bs x}_{t-1} =  \bs{\mu}_{\bs{\theta}}({\bs x}_t, t)  + {\bs \Sigma }^{1/2}_{\bs \theta}( {\bs x}_t, t ) {\bs z} + w {\bs \Sigma}_{\bs \theta} ( {\bs x}_t, t ) {\bs g}^*({\bs x_t}, t; s).
\end{align}
See~\cref{alg:sampling} for pseudocode of our sampler that incorporates the intermittent technique. The generation due to~\cref{eq:guided_ours} can be interpreted as sampling from a modified density $\tilde{p}_{\bs \theta} ( {\bs x}_t ) \propto p_{\bs \theta} ( {\bs x}_t ) e^{ \tilde{{\cal L}} ( {\bs x}_t; s ) } $. We note that instances with high values of $\tilde{{\cal L}} ( {\bs x}_t; s )$ would have more chances to be generated compared to the original density $p_{\bs \theta} ( {\bs x}_t )$, which aligns well with our focus to encourage the generation of minorities.


\section{Experiments}


\subsection{Setup}

\textbf{Datasets and pretrained models.} We employ four real-world benchmarks that include both unconditional and conditional data. Our unconditional benchmarks are CelebA $64^2$~\cite{liu2015faceattributes} and LSUN-Bedrooms $256^2$~\cite{yu2015lsun}. For the class-conditional datasets, we employ ImageNet $64^2$ and $256^2$~\cite{deng2009imagenet}. The pretrained model for CelebA was constructed by ourselves by following the settings in~\cite{um2023don}. The models for LSUN-Bedrooms and ImageNet were taken from the checkpoints provided in~\cite{dhariwal2021diffusion}. In addition to our primary focus on these four benchmarks, we further explore challenging scenarios of T2I generation and medical imaging to scrutinize the boundaries of our approach. See the supplementary for explicit details on these experimental tasks.

\begin{figure}[!t]
    \begin{subfigure}[h]{0.321\textwidth}
    \centering
    \includegraphics[width=1.0\columnwidth]{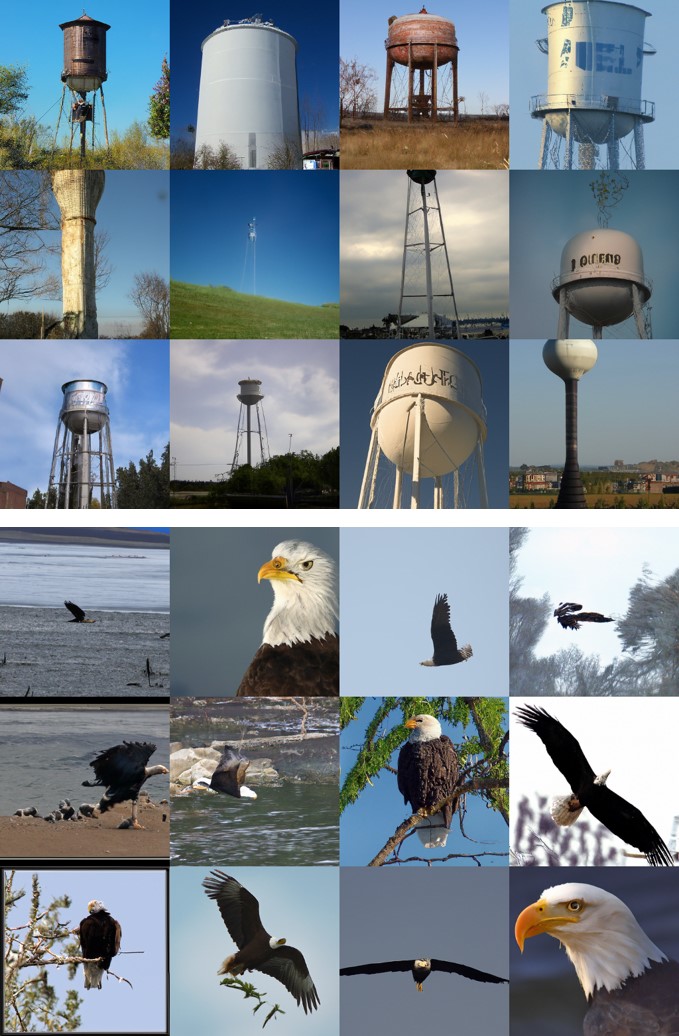}
    \caption{ADM~\cite{dhariwal2021diffusion}}
    \end{subfigure}
    \hfill
    \begin{subfigure}[h]{0.321\textwidth}
    \centering
    \includegraphics[width=1.0\columnwidth]{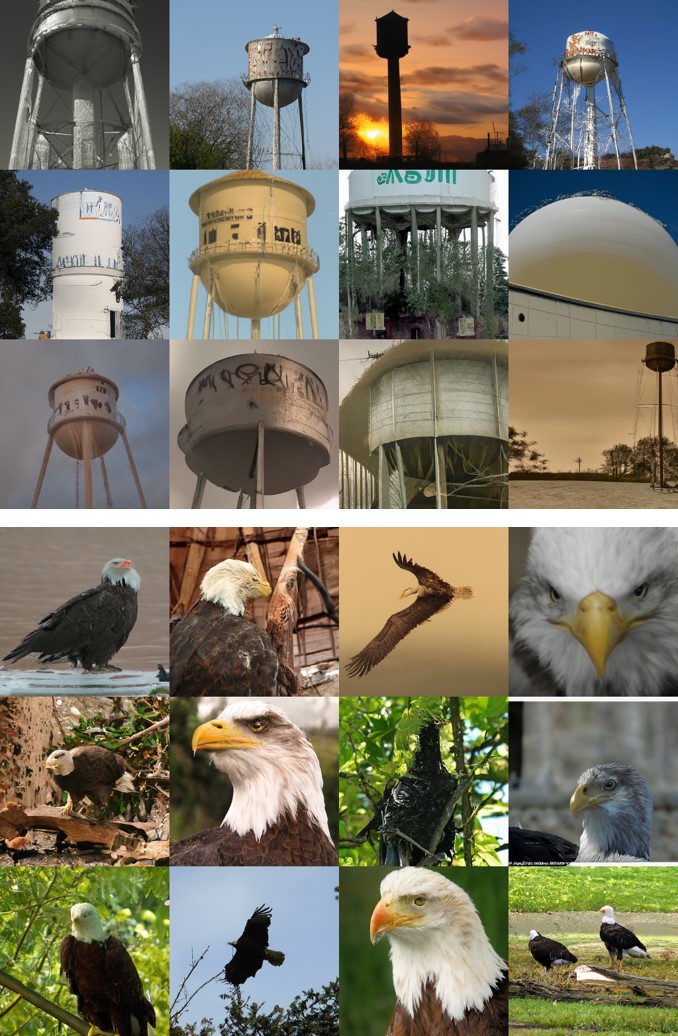}
    \caption{Sehwag \etal~\cite{sehwag2022generating}}
    \end{subfigure}
    \hfill
    \begin{subfigure}[h]{0.321\textwidth}
    \centering
    \includegraphics[width=1.0\columnwidth]{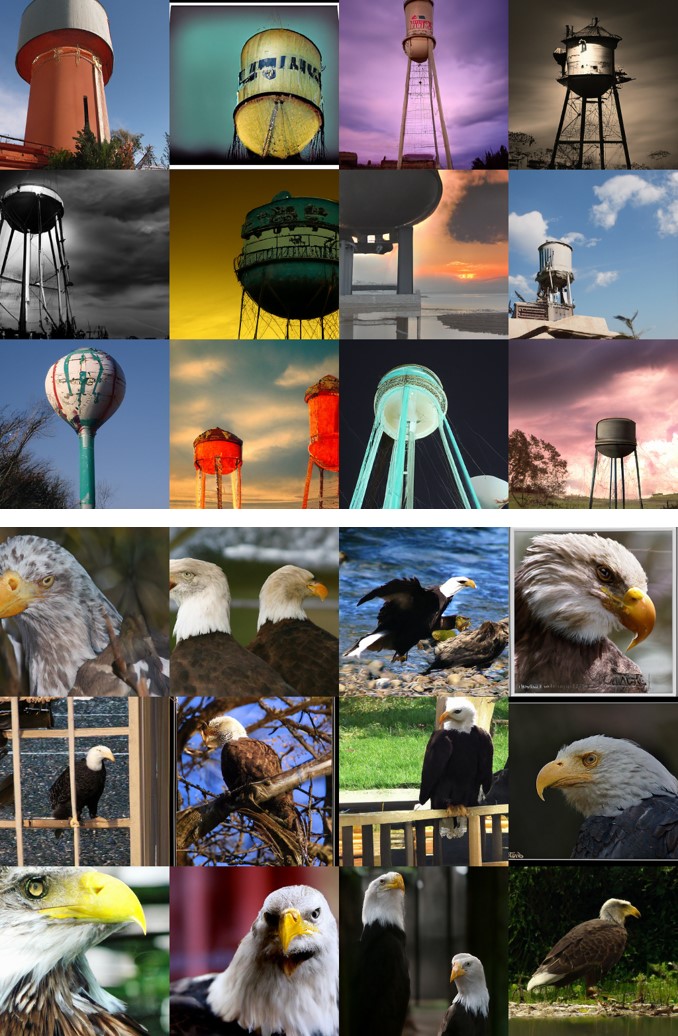}
    \caption{Ours}
    \end{subfigure}
\caption{Sample comparison on ImageNet-256. Generated samples from two classes are exhibited: Water tower (top row) and Bald eagle (bottom row). For each row, we share the same random seed across all three methods.}
\label{fig:imagenet_samples}
\end{figure}

\noindent \textbf{Baselines.} We compare a variety of frameworks with our approach, which encompasses existing minority samplers as well as generic frameworks that are not specifically tailored for low-density generation. For the CelebA experiments, we employ BigGAN~\cite{brock2018large}, ADM~\cite{dhariwal2021diffusion} with ancestral sampling, and Um and Ye~\cite{um2023don}. As in~\cite{um2023don}, we incorporate an additional baseline on CelebA, which implements a conditioned generation of minority instances by using the classifier guidance~\cite{dhariwal2021diffusion} with minority annotations given in the dataset (\eg, ``Eyeglasses''~\cite{yu2020inclusive}). We also compare~\cite{sehwag2022generating} on CelebA by extending their sampler to admit unconditional data, where we used the same approach as~\cite{um2023don} for the extension. For LSUN-Bedrooms, we consider four baselines: (i) StyleGAN~\cite{karras2019style}; (ii) ADM~\cite{dhariwal2021diffusion}; (iii) LDM~\cite{rombach2022high}; (iv) Um and Ye~\cite{um2023don}. We focus on four diffusion-based frameworks on ImageNet-64: (i) ADM~\cite{dhariwal2021diffusion}; (ii) EDM~\cite{karras2022elucidating}; (iii) Sehwag \etal~\cite{sehwag2022generating}; (iv) Um and Ye~\cite{um2023don}. For the ImageNet-256 experiments, we consider: (i) ADM~\cite{dhariwal2021diffusion}; (ii) DiT~\cite{Peebles2022DiT}; (iii) CADS~\cite{sadat2023cads}; (iv) Sehwag \etal~\cite{sehwag2022generating}; (v) Um and Ye~\cite{um2023don}.

\begin{table}[t]
\caption{Comparison of sample quality and diversity. ``ADM-ML'' refers to a classifier-guided sampler implemented on ADM~\cite{dhariwal2021diffusion}, which conditions on \textbf{M}inority \textbf{L}abels~\cite{um2023don}. ``+ intermittent'' refers to our sampler that incorporates intermittent guidance. For baseline real data, we employ the most unique samples that yield the highest AvgkNN values. The best results are marked in \textbf{bold}, and the second bests are \underline{underlined}.}
\label{table:metrics}
\fontsize{9}{9}\selectfont{
    \begin{subtable}[h]{0.499\textwidth}
        \centering
        \begin{tabular}{lcccc}
            \toprule[0.1em]
              \multicolumn{1}{l}{Method}  & \multicolumn{1}{c}{cFID} & \multicolumn{1}{c}{sFID} & \multicolumn{1}{c}{Prec} & \multicolumn{1}{c}{Rec} \\
              \\
              \multicolumn{5}{l}{\textbf{CelebA 64$\times$64}}\\
            \toprule[0.1em]
            \multicolumn{1}{l}{ADM~\cite{dhariwal2021diffusion}} & 75.41  & 17.11 & \textbf{0.97} & 0.23 \\
            \multicolumn{1}{l}{BigGAN~\cite{brock2018large}} & 80.58 & 16.80 & \textbf{0.97} & 0.19 \\
            \midrule[0.001em]
            \multicolumn{1}{l}{ADM-ML~\cite{um2023don}} & 51.99 & 13.40 & \underline{0.94} & 0.30 \\
            \multicolumn{1}{l}{Sehwag \etal \cite{sehwag2022generating}} & 28.25 & 10.64 & 0.82 & 0.42 \\
            \multicolumn{1}{l}{Um and Ye \cite{um2023don}} & 27.32 & 8.66  & 0.89 & 0.33 \\
            \midrule[0.001em]
            \multicolumn{1}{l}{Ours} & \textbf{18.57} & \textbf{8.20} & 0.83 & \textbf{0.48} \\
            \multicolumn{1}{l}{+ intermittent} & \underline{19.34} & \underline{8.85} & 0.82 & \underline{0.47} \\
            \\
            \multicolumn{5}{l}{\textbf{ImageNet 64$\times$64}}\\
            \toprule[0.1em]
            \multicolumn{1}{l}{ADM~\cite{dhariwal2021diffusion}} & 18.37 & 5.39 & \underline{0.79} & 0.53 \\
            \multicolumn{1}{l}{EDM~\cite{karras2022elucidating}} & 19.09 & 4.73 & 0.73 & 0.59 \\
            \midrule[0.001em]
            \multicolumn{1}{l}{Sehwag \etal \cite{sehwag2022generating}} & 11.37 & 4.69 & \textbf{0.80} & 0.52 \\
            \multicolumn{1}{l}{Um and Ye \cite{um2023don}}  & 12.47 & \underline{3.13} & 0.76 & 0.56 \\
            \midrule[0.001em]
            \multicolumn{1}{l}{Ours} & \textbf{11.08} & \textbf{3.09} & 0.72 & \textbf{0.63} \\
            \multicolumn{1}{l}{+ intermittent} & \underline{11.24} & 3.17 & 0.73 & \underline{0.62} \\
            \\
        \end{tabular}
    \end{subtable}
    \begin{subtable}[h]{0.499\textwidth}
        \centering
        \begin{tabular}{lcccc}
            \toprule[0.1em]
              \multicolumn{1}{l}{Method}  & \multicolumn{1}{c}{cFID} & \multicolumn{1}{c}{sFID} & \multicolumn{1}{c}{Prec} & \multicolumn{1}{c}{Rec} \\
              \\
            \multicolumn{5}{l}{\textbf{LSUN Bedrooms 256$\times$256}}\\
            \toprule[0.1em]
            \multicolumn{1}{l}{ADM~\cite{dhariwal2021diffusion}} & 63.30 & 8.00 & \underline{0.89} & \underline{0.15} \\
            \multicolumn{1}{l}{LDM~\cite{rombach2022high}} & 63.53 & 7.73 & \textbf{0.90} & 0.13 \\
            \multicolumn{1}{l}{StyleGAN~\cite{karras2019style}} & 57.17 & 7.78 & \underline{0.89} & 0.14 \\
            \midrule[0.001em]
            \multicolumn{1}{l}{Um and Ye \cite{um2023don}} & 41.75 & 7.26 & 0.87 & 0.10 \\
            \midrule[0.001em]
            \multicolumn{1}{l}{Ours} & \textbf{35.79} & \textbf{4.94} & 0.86 & \textbf{0.16} \\
            \multicolumn{1}{l}{+ intermittent} & \underline{36.94} & \underline{5.13} & 0.87 & \underline{0.15} \\
            \\
            \multicolumn{5}{l}{\textbf{ImageNet 256$\times$256}}\\
            \toprule[0.1em]
            \multicolumn{1}{l}{ADM~\cite{dhariwal2021diffusion}} & 13.22 & 7.66 & \textbf{0.86} & 0.39 \\
            \multicolumn{1}{l}{DiT~\cite{Peebles2022DiT}} & 21.51 & 6.76 & 0.80 & \underline{0.46} \\
            \multicolumn{1}{l}{CADS~\cite{sadat2023cads}} & 15.95 & 6.18 & 0.81 & \textbf{0.48} \\
            \midrule[0.001em]
            \multicolumn{1}{l}{Sehwag \etal \cite{sehwag2022generating}} & 10.93 & 6.66 & \underline{0.85} & 0.39 \\
            \multicolumn{1}{l}{Um and Ye \cite{um2023don}}  & 11.44 & 4.63 & \underline{0.85} & 0.42 \\
            \midrule[0.001em]
            \multicolumn{1}{l}{Ours} & \textbf{9.93} & \textbf{4.19} & 0.83 & \underline{0.46} \\
            \multicolumn{1}{l}{+ intermittent} & \underline{9.98} & \underline{4.35} & 0.83 & 0.45 \\
            \\
        \end{tabular}
    \end{subtable}
    }
\end{table}

\noindent \textbf{Evaluation metrics.} We respect the choices in the previous studies~\cite{sehwag2022generating, um2023don} and employ three distinct measures for evaluating low-densitiness of the considered methods: (i) Average k-Nearest Neighbor (AvgkNN); (ii) Local Outlier Factor (LOF)~\cite{breunig2000lof}; (iii) Rarity Score~\cite{han2022rarity}. For all three measures, a higher value implies that the instance is less likely than its neighborhood samples~\cite{sehwag2022generating, han2022rarity, um2023don}. We also employ a range of quantitative metrics to assessing quality and diversity, including: (i) Clean Fr\'echet Inception Distance (cFID)~\cite{parmar2022aliased}; (ii) Spatial FID (sFID)~\cite{nash2021generating}; (iii) Improved Precision \& Recall~\cite{kynkaanniemi2019improved}. Specifically to evaluate the proximity to real minority data, we follow the same approach as in~\cite{um2023don} and employ instances with the lowest likelihoods (\eg, the ones yielding the highest AvgkNNs) as baseline real data for calculating our quality and diversity metrics.


\subsection{Results}

\noindent {\bf Qualitative comparisons.} \cref{fig:lsun_samples} compares generated samples on the LSUN-Bedrooms dataset. Notice that both minority generators (\ie, Um and Ye~\cite{um2023don} and ours) are more likely to yield low-likelihood features of the dataset (\eg, complex visual attributes~\cite{serra2019input, arvinte2023investigating}) compared to a standard ancestral sampling implemented with ADM~\cite{dhariwal2021diffusion}. An important distinction herein is that our method yields this performance benefit solely with a pretrained model, in contrast to~\cite{um2023don} that requires significant resources to train a separate classifier.~\cref{fig:imagenet_samples} exhibits generated samples on another challenging benchmark, ImageNet $256 \times 256$. We see similar benefits of our method compared to the baselines, further demonstrating the effectiveness of our sampler on challenging large-scale benchmarks. See the supplementary for samples on CelebA.

\noindent \textbf{Quantitative evaluation.} \cref{table:metrics} exhibits quality and diversity evaluations on our focused benchmarks. For the baseline real data, we employ the most unique samples that yield the highest AvgkNN values. Notice that our sampler yields better (or comparable) results than the baseline approaches in all datasets, demonstrating its ability to produce high-quality diverse samples close to the baseline real minorities. We emphasize that this superior performance persists even with the intermittent technique that can significantly reduce our inference time to a similar extent as existing samplers; see the supplementary for a detailed complexity analysis. We further highlight that the benefit of ours comes solely from a pretrained model, implying the practical importance of our approach.

\begin{figure}[!t]
\begin{center}
\includegraphics[width=1.0\columnwidth]{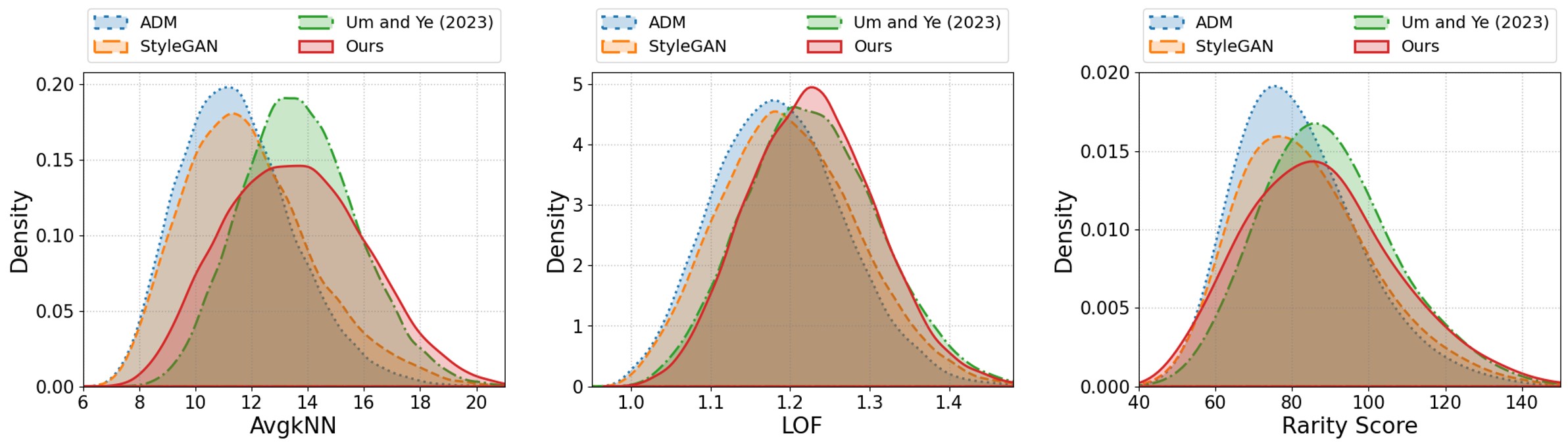}
\end{center}
\caption{Comparison of neighborhood density on LSUN-Bedrooms. “AvgkNN” refers to Average k-Nearest Neighbor, and “LOF” is Local Outlier Factor~\cite{breunig2000lof}. ``Rarity Score'' indicates a low-density metric proposed by~\cite{han2022rarity}. The higher values, the less likely samples for all three measures.}
\label{fig:lsun_densities}
\end{figure}

\noindent \textbf{Neighborhood density results.} \cref{fig:lsun_densities} compares our focused density measures on LSUN-Bedrooms. Observe that for all three metrics, both our method and~\cite{um2023don} greatly improve the capability of generating minorities over ancestral sampling (implemented with ADM~\cite{dhariwal2021diffusion}), which corroborates the visual inspections made on~\cref{fig:lsun_samples}. However, we highlight that our results were attained only with a pretrained model, which is in contrast to~\cite{um2023don} helped by an external classifier to yield the benefit. See the supplementary for density results on other datasets.

\begin{table}[t!]
    \caption{Ablation study investigating various design elements in our sampler. ``Cost'' denotes inference time measured in sec/sample. ``\xmark'' refers to ours not incorporating the stop gradient. ``$\texttt{sg}(\hat{\bs x}_0)$'' and ``$\texttt{sg}( \doublehat{{\bs x}}_0 )$'' indicates the cases with the stop gradient either on $\hat{\bs x}_0$ and $\doublehat{{\bs x}}_0$, respectively. ``S/O'' denotes the case that uses the sudden switch-off: $w_t = w \cdot \mathbbm{1}\{t \ge t_{\text{mid}} \}$. ``Var'' refers to ours using the variance schedule: $w_t = w \cdot {\bs \Sigma}_{\bs \theta} ( {\bs x}_t, t )$.}
    \label{table:ablations}
    \centering
    \begin{subtable}[t]{0.325\textwidth}
        \caption{Influence of the scale \(w\)}
        \label{table:impact_w}
        \centering
        \scalebox{0.85}{
            \begin{booktabs}{colspec = {Q[c, 0.5cm]Q[c, 1.0cm]Q[c, 0.8cm]}, row{1-Z}={font=\footnotesize}}
                \toprule
                \(w\) & cFID $\downarrow$ & Rec $\uparrow$ \\
                \midrule
                0.0 & 84.51  & 0.18 \\
                2.0 & 57.98 & 0.56 \\
                6.0 & \textbf{34.01} & \textbf{0.63} \\
                \bottomrule
            \end{booktabs}
        }
    \end{subtable}
    \begin{subtable}[t]{0.325\textwidth}
        \caption{Impact of using $\texttt{sg}(\cdot)$}
        \label{table:impact_sg}
        \centering
        \scalebox{0.85}{
            \begin{booktabs}{colspec = {Q[c, 1.0cm]Q[c, 1.0cm]Q[c, 1.0cm]}, row{1-Z}={font=\footnotesize}}
                \toprule
                Type & cFID $\downarrow$ & Cost $\downarrow$ \\
                \midrule
                \xmark & \textbf{32.09}  &  2.65 \\
                $\texttt{sg}( \hat{{\bs x}}_0 )$ & 81.08 &  2.62 \\
                $\texttt{sg}( \doublehat{{\bs x}}_0 )$ & 34.01 &  \textbf{1.75} \\
                \bottomrule
            \end{booktabs}
        }
    \end{subtable}
    \begin{subtable}[t]{0.325\textwidth}
        \caption{Effect of the schedule $w_t$}
        \label{table:impact_w_t}
        \centering
        \scalebox{0.85}{
            \begin{booktabs}{colspec = {Q[c, 1.0cm]Q[c, 1.0cm]Q[c, 0.9cm]}, row{1-Z}={font=\footnotesize}}
                \toprule
                Type & cFID $\downarrow$ & Prec $\uparrow$  \\
                \midrule
                Fixed  &  34.01  & 0.56 \\
                S/O  & \textbf{31.79} & 0.57  \\
                Var  & 37.78 & \textbf{0.87}  \\
                \bottomrule
            \end{booktabs}
        }
    \end{subtable}
\end{table}

\noindent {\bf Ablation studies.}~\cref{table:ablations} exhibits ablations on our important design choices. Notice in~\cref{table:impact_w} that increasing $w$ yields improvements in proximity to real minority instances, validating the role of $w$ as a knob to control the strength of our guidance.~\cref{table:impact_sg} shows the benefit of incorporating the stop-gradient on $\doublehat{{\bs x}}_0$, where we see noticeable gain in the inference time with little compromise in performance.~\cref{table:impact_w_t} investigates the impact of time-scheduling strategies. Note that the variance-based scheduling demonstrates significant improvement in sample quality while maintaining respectable cFID performance, providing justification for adopting the variance schedule as our final scheduling strategy. See the supplementary for further analyses and ablations on other parameters.

\begin{wraptable}{r}{0.5\textwidth}
    \vspace{-0.9cm}
    \caption{Classification results for different training datasets. All settings were evaluated on the CelebA testset and averaged over three distinct runs.}
    \label{tab:classification}
    \centering
    \fontsize{8}{8}\selectfont
        {
        \scalebox{0.9}{
    \begin{booktabs}{colspec = {Q[l, 2.9cm]Q[c, 0.65cm]Q[c, 0.65cm]Q[c, 0.65cm]}}
        \toprule
          Training data  & F1 & Prec & Rec \\
        \midrule
         CelebA trainset & 0.746  &  0.815  & 0.710  \\
         + ADM gens (50K)   & 0.742  & 0.808  & 0.711  \\
        + Ours gens (50K)   & \textbf{0.757} & \textbf{0.822} & \textbf{0.724} \\
        \bottomrule
    \end{booktabs}
    }}
    \vspace{-0.6cm}
\end{wraptable}
\noindent {\bf Downstream classification.} To further emphasize the practical significance of our work, we explore a potential application of our sampler. Specifically, we investigate whether our minority-enhanced generated samples could enhance the performance of a classifier trained on a synthetically augmented dataset. We consider the classification of 40 distinct attributes of CelebA and train ResNet-18 models on three different datasets: (i) CelebA trainset; (ii) CelebA + 50K samples from ADM~\cite{dhariwal2021diffusion}; (iii) CelebA + 50K samples from ours. We incorporated an off-the-shelf classifier for labeling the generated samples.~\cref{tab:classification} compares the classification metrics of the three considered cases. Note that the ADM-augmented classifier fails to improve upon the non-augmented case, which we conjecture is due to the limited diversity of the ADM samples, a factor that has been observed to potentially hinder performance improvement~\cite{ravuri2019classification, gowal2021improving, zhao2022synthesizing}. Nonetheless, the classifier complemented with our samples enhances the performance across all metrics, highlighting the benefit of ours for downstream applications.


\section{Conclusion}
\label{sec:conclusion}

We develop a novel framework for generating minority data using diffusion models. Our self-guided sampler, based on our new minority metric, optimizes the generation process of diffusion models to evolve towards low-likelihood minority features. We additionally provide several techniques to further improve the complexity and fidelity due to our sampler. Extensive experiments across real data benchmarks demonstrate significant improvements over existing minority samplers. Importantly, the benefits stem solely from a pretrained diffusion model, distinguishing our approach from existing frameworks requiring additional components to improve the minority-generating capability over standard samplers.

\noindent {\bf Limitation and potential negative impact.} One disadvantage is that the proposed sampler introduces additional inference costs compared to standard samplers. A potential concern is the misuse of our sampler to intentionally suppress the generation of minority-featured samples. This malicious use could be realized by employing negative values of $w$ in~\cref{eq:guided_ours}, directing the focus towards producing instances dominated by high-likelihood majority features. It is crucial to acknowledge and address this risk, emphasizing the need for responsible usage of our framework to uphold fairness and inclusivity in generative modeling.

\section*{Acknowledgments} 

This work was partly supported by the National Research Foundation of Korea under Grant (No. RS-2024-00336454), by Institute of Information \& communications Technology Planning \& Evaluation (IITP) grant funded by the Korea government (MSIT)) (No. RS-2022-II220984, Development of Artificial Intelligence Technology for Personalized Plug-and-Play Explanation and Verification of Explanation), by Institute for Information \&
communications Technology Promotion (IITP) grant funded by the Korea government (MSIT) 
(No. RS-2019-II190075 Artificial Intelligence Graduate School Program (KAIST)), by Culture, Sports and Tourism R\&D Program through the Korea Creative Content Agency grant funded by the Ministry of Culture, Sports and Tourism in 2023, and by Field-oriented Technology Development Project for Customs Administration funded by the Korea government (the Ministry of Science \& ICT and the Korea Customs Service) through the National Research Foundation (NRF) of Korea under Grant NRF2021M3I1A1097910.

%
%
\bibliographystyle{splncs04}
\bibliography{main}

\clearpage

\beginsupplement

\section{Proofs}

\subsection{Proof of~\cref{proposition}}

\begin{manualproposition}{\ref{proposition}}
    Consider minority score in~\cref{eq:ms} with the squared-error distance loss $\| \cdot \|_2^2$. Its weighted sum over timesteps is equivalent (upto a constant factor) to the negative ELBO considered in~\cite{ho2020denoising}:
    \begin{align*}
        \sum_{t=1}^T \bar{w}_t {\cal L}({\bs x}_0; t)
         = \sum_{t=1}^T {\mathbb E}_{p({\bs \epsilon})}
        [ \| {\bs \epsilon} -  {\bs \epsilon}_{\bs \theta} ( \sqrt{\alpha_t} {\bs x}_0 + \sqrt{1 - \alpha_t} {\bs \epsilon}, t )  \|_2^2  ] \gtrapprox - \log p_{\bs \theta} ({\bs x}_0),
    \end{align*}
    where $\bar{w}_t \coloneqq \alpha_t / (1- \alpha_t) $ and $p ( {\bs \epsilon} ) \coloneqq {\cal N}({\bs \epsilon} ; {\bs 0}, {\bs I})$.
\end{manualproposition}
\begin{proof}
We start from the definition of minority score in~\cref{eq:ms}:
\begin{align*}
    {\cal L}({\bs x}_0; t) \coloneqq 
    \mathbb{E}_{q_{\alpha_t} ({\bs x}_t \mid {\bs x}_0) } [ d( {\bs x}_0, \hat{{\bs x}}_0 ({\bs x}_t) ) ].
\end{align*}
Plugging the squared-error loss and further manipulations then yield:
\begin{align}
\label{eq:ms_to_noisematching}
    {\cal L}({\bs x}_0; t) \coloneqq  \; &
    \mathbb{E}_{q_{\alpha_t} ({\bs x}_t \mid {\bs x}_0) } [ d({\bs x}_0, \hat{{\bs x}}_0 ({\bs x}_t) ) ] 
    =  \mathbb{E}_{q_{\alpha_t} ({\bs x}_t \mid {\bs x}_0) } [ \| {\bs x}_0 - \hat{{\bs x}}_0  \|_2^2 ] \nonumber \\
    = \; & \mathbb{E}_{q_{\alpha_t} ({\bs x}_t \mid {\bs x}_0) } \left[   \left\|    {\bs x}_0 -  \frac{1}{\sqrt{\alpha_t}} \{    {\bs x}_t -  \sqrt{1-\alpha_t} {\bs \epsilon}_{\bs \theta}  ({\bs x}_t, t) \}             \right\|_2^2   \right] \nonumber \\
    = \; & \mathbb{E}_{p ({\bs \epsilon}) }  \left[   \left\|  \frac{ \sqrt{1-\alpha_t} }{ \sqrt{\alpha_t} }      \{  {\bs \epsilon} -  {\bs \epsilon_{\theta}  }({\bs x}_t, t)  \}           \right\|_2^2   \right] \nonumber \\ 
    = \; & \frac{1 - \alpha_t}{\alpha_t}
    \mathbb{E}_{p ({\bs \epsilon}) }  \left[ \| {\bs \epsilon} - {\bs \epsilon_{\theta}  }({\bs x}_t, t)   \|_2^2 \right] \nonumber \\
    = \; & \tilde{w}_t \mathbb{E}_{p ({\bs \epsilon}) }  \left[ \|  {\bs \epsilon} -  {\bs \epsilon}_{\bs \theta} ( \sqrt{\alpha_t} {\bs x}_0 + \sqrt{1 - \alpha_t} {\bs \epsilon}, t ) \|_2^2 \right],
\end{align}
where $\tilde{w}_t \coloneqq (1-\alpha_t) / \alpha_t$. The second equality is due to the Tweedie's formula (\ie, \cref{eq:tweedie_with_score}) together with the noise-predicting expression ${\bs s}_{\bs{\theta}}( {\bs x}_t, t )  =  -  {\bs \epsilon_{\theta}}({\bs x}_t, t) / \sqrt{1 - \alpha_t} $. A weighted sum of the expression in~\cref{eq:ms_to_noisematching} over timesteps with $\bar{w}_t \coloneqq 1 / \tilde{w}_t = \alpha_t / (1- \alpha_t) $ gives:
\begin{align*}
    \sum_{t=1}^T \bar{w}_t  {\cal L}({\bs x}_0; t)   =   \sum_{t=1}^T \mathbb{E}_{p ({\bs \epsilon}) }  \left[ \|  {\bs \epsilon} -  {\bs \epsilon}_{\bs \theta} ( \sqrt{\alpha_t} {\bs x}_0 + \sqrt{1 - \alpha_t} {\bs \epsilon}, t )  \|_2^2 \right].
\end{align*}
Notice that the RHS is equivalent (up to a constant) to the expression of the negative ELBO considered in DDPM~\cite{ho2020denoising, li2023your}. This completes the proof.
\end{proof}

\subsection{Proof of~\cref{corollary}}

\begin{manualcorollary}{\ref{corollary}}
    The proposed metric in~\cref{eq:mse} with the squared-error loss is equivalent to the negative ELBO w.r.t. $\log p_{\bs \theta} (\hat{{\bs x}}_0 ({\bs x}_t))$ when integrated over timesteps with $\bar{w}_s \coloneqq \alpha_s / (1- \alpha_s)$.
\end{manualcorollary}
\begin{proof}
    The proof is immediate with~\cref{proposition} and the relation between the two minority metrics. Since our metric is interpretable as minority score of $\hat{{\bs x}}_0({\bs x}_t)$, we have
    \begin{align*}
        \tilde{{\cal L}} ( {\bs x}_t; s ) = {\cal L} ( \hat{{\bs x}}_0 ({\bs x}_t); s ).
    \end{align*}
    Integrating over timesteps with $\bar{w}_s \coloneqq \alpha_s / (1- \alpha_s)$ gives:
    \begin{align*}
        \sum_{s=1}^T \bar{w}_s \tilde{{\cal L}} ( {\bs x}_t; s ) & = \sum_{s=1}^T \bar{w}_s {\cal L} ( \hat{{\bs x}}_0 ({\bs x}_t); s ) \\
        & = \sum_{s=1}^T {\mathbb E}_{p({\bs \epsilon})}
        [ \| {\bs \epsilon} -  {\bs \epsilon}_{\bs \theta} ( \sqrt{\alpha_s} \hat{{\bs x}}_0 ( {\bs x}_s ) + \sqrt{1 - \alpha_s} {\bs \epsilon}, s )  \|_2^2  ],
    \end{align*}
    where the second equality follows from~\cref{proposition}. Note that the RHS of the second equality is equivalent (up to a constant) to the expression of the negative ELBO w.r.t. $\hat{{\bs x}}_0 ({\bs x}_t)$. This completes the proof.
\end{proof}

\section{Additional Details on Experimental Setup}
\label{sec:imple}

\noindent \textbf{Pretrained models.} The pretrained model for CelebA was constructed by ourselves by respecting the settings in~\cite{um2023don}. The models for LSUN-Bedrooms and ImageNet were taken from the checkpoints provided in~\cite{dhariwal2021diffusion}. As in~\cite{sehwag2022generating}, we leveraged the upscaling model developed in~\cite{dhariwal2021diffusion} for the results on ImageNet-256.

\noindent \textbf{Baselines.} The ADM~\cite{dhariwal2021diffusion} baselines on the four main benchmarks leveraged the same pretrained models as our approach. For implementing the sampler due to~\cite{um2023don}, we respected the settings provided in their manuscript for all considered datasets. Specifically based on the ADM pretrained models (\ie, the same ones as ours), we employed encoder architectures of U-Net for minority classifiers and incorporated all training samples to construct the classifiers, except for the one on LSUN-Bedrooms where only a 10\% of the training set were used. For the ImageNet-256 results of~\cite{um2023don}, we employed the upscaling model~\cite{dhariwal2021diffusion} as in~\cite{um2023don}.

The BigGAN model for our CelebA experiments is based on the same architecture used in~\cite{choi2020fair}\footnote{\url{https://github.com/ermongroup/fairgen}}, and we respect the training setup provided in the official project page of BigGAN\footnote{\url{https://github.com/ajbrock/BigGAN-PyTorch}}. For the additional baseline on CelebA with the classifier guidance targeting minority annotations (\ie, ADM-ML in~\cref{table:metrics}), the classifier was trained to predict four minority attributes: (i) ``Bald''; (ii) ``Eyeglasses''; (iii) ``Mustache''; (iv) ``Wearing\_Hat''. During inference time, we generated samples with random combinations of the four attributes (\eg, bald hair yet not wearing glasses) using the classifier guidance. The backbone model used for ADM-ML is the same as ours. To implement the sampler by~\cite{sehwag2022generating} on CelebA, we constructed an out-of-distribution (OOD) classifier that predicts whether a given input is from CelebA or other datasets (\eg, ImageNet). We then incorporated the gradient of negative log-likelihood of the classifier (targeting the in-distribution class) into ancestral sampling to yield low-likelihood guidance, complemented by a real-fake discriminator to enhance sample quality (as proposed in~\cite{sehwag2022generating}). For the implementation of~\cite{um2023don} on CelebA, we used the same pretrained model as ours and respected the same hyperparameter setup as in their original paper.

For the LDM~\cite{rombach2022high} baseline on LSUN-Bedrooms, we employed the checkpoint provided by~\cite{rombach2022high}\footnote{\url{https://github.com/CompVis/latent-diffusion}}. The StyleGAN results were obtained via the pretrained model offered by the authors~\cite{karras2019style}\footnote{\url{https://github.com/NVlabs/stylegan}}. For~\cite{um2023don} on LSUN-Bedrooms, we leveraged the same pretrained model as our sampler (\ie, ADM) with the original hyperparameters reported in the paper~\cite{um2023don}.

The EDM~\cite{karras2022elucidating} baseline on ImageNet-64 employed the checkpoint provided in the official project page of~\cite{karras2022elucidating}\footnote{\url{https://github.com/NVlabs/edm}}. For~\cite{sehwag2022generating} on ImageNet-64, we employed the pretrained classifier provided by~\cite{dhariwal2021diffusion} and constructed a discriminator by following the setting described in~\cite{sehwag2022generating}. The pretrained model was the same as ours.

The DiT~\cite{Peebles2022DiT} baseline on ImageNet-256 employed the pretrained checkpoint provided in the code repository provided by the authors\footnote{\url{https://github.com/facebookresearch/DiT}}. The official codebase of CADS~\cite{sadat2023cads} was not publicly available, so we resorted to our own implementation to yield the results, which is based on the pseudocode provided in the manuscript~\cite{sadat2023cads}. We employed the hyperparameter settings recommended in the original manuscript~\cite{sadat2023cads}. For~\cite{sehwag2022generating} on ImageNet-256, we employed the upscaling model by following the original setting taken in~\cite{sehwag2022generating}.

\noindent \textbf{Evaluation metrics.} To compute Local Outlier Factor (LOF)~\cite{breunig2000lof} of generated samples, we employed the implementation in PyOD~\cite{zhao2019pyod}\footnote{\url{https://pyod.readthedocs.io/en/latest/}}. The numbers of nearest neighbors for computing AvgkNN and LOF were chosen as 5 and 20, respectively, which are the conventional values widely used in practice. As in~\cite{sehwag2022generating, um2023don}, AvgkNN and LOF were computed in the feature space of ResNet-50. We computed Rarity Score~\cite{han2022rarity} with $k=5$ using implementation provided in the official project page\footnote{\url{https://github.com/hichoe95/Rarity-Score}}.

Clean Fr\'echet Inception Distance (cFID)~\cite{parmar2022aliased} were evaluated via the official implementation\footnote{\url{https://github.com/GaParmar/clean-fid}}. We evaluated spatial FID~\cite{nash2021generating} based on the official pytorch FID~\cite{heusel2017gans}\footnote{\url{https://github.com/mseitzer/pytorch-fid}} with some modifications to leverage spatial features (\ie, the first 7 channels from the intermediate \texttt{mixed\_6/conv} feature maps), instead of using the standard \texttt{pool\_3} inception features. The results of Improved Precision \& Recall~\cite{kynkaanniemi2019improved} were obtained with $k=5$ using the official codebase of~\cite{han2022rarity}. To evaluate the closeness to low-likelihood instances residing on tail of data, we employed the least probable instances as baseline real data for computing the quality metrics. Specifically we used the 10K real samples yielding the highest AvgkNN values for CelebA. For LSUN-Bedrooms and ImageNet, the most unique 50K samples that yield the highest AvgkNNs were employed for baseline real data. All quality and diversity metrics were computed with 30K generated samples.

\noindent \textbf{Hyperparameters.} For the discrepancy notion $d(\cdot, \cdot)$ in the proposed metric (\ie,~\cref{eq:mse}), we employed LPIPS~\cite{zhang2018perceptual} for our four main benchmarks\footnote{One may concern that the use of LPIPS in our framework seems conflicting with our claimed benefit of the self-containment, since the metric incorporates an external neural network (like AlexNet~\cite{krizhevsky2012imagenet}) for computing perceptual loss. Nonetheless, we argue that our reliance on LPIPS stands apart from the external dependencies observed in prior works, which often involve time-consuming procedures like training a classifier~\cite{sehwag2022generating, um2023don}. Notably, our reliance on LPIPS is nearly cost-free, as it is readily accessible through the open-source package.}. For perturbation timestep $s$, we used $0.8T$ for the cosine-scheduled~\cite{nichol2021improved} models (\ie, CelebA and ImageNet), while $0.5T$ for the linear-scheduled~\cite{ho2020denoising} diffusion model (like the one used on LSUN-Bedrooms). For the number of Monte-Carlo samples drawn from $q_{\alpha_s} ( \hat{\bs x}_s | \hat{\bs x}_0  )$ in~\cref{eq:mse} (to approximate the expectation), we employed only a single sample globally.

In line with~\cite{sehwag2022generating}, we incorporated a normalization technique into our guidance term to ensure a unit $l_{\infty}$ norm. More precisely, we used $\tilde{{\bs g}}^*( {\bs x}_t, t;s) \coloneqq  {\bs g}^*({\bs x_t}, t;s) / \|{\bs g}^*({\bs x_t}, t;s) \|_{\infty} $ instead of ${\bs g}^*({\bs x_t}, t;s)$. We leveraged the variance schedule (\ie, $w_t = w{\bs \Sigma}_{\bs \theta} ( {\bs x}_t, t )$) for all of our datasets. For the scale constant $w$, we used $0.4$ and $0.25$ for CelebA and LSUN-Bedrooms, respectively. On the other hand, we employed $w=0.2$ for the ImageNet results. For the intermittent rate $n$, we used $n=5$ on CelebA and LSUN-Bedrooms while employing $n=2$ for the ImageNet experiments. On ImageNet-256, our guidance was incorporated during generations on $64 \times 64$ samples. The subsequent upscaling to $256 \times 256$ was carried out using ancestral sampling, following the same approach as~\cite{sehwag2022generating, um2023don}. We globally employed 250 timesteps to sample from all diffusion-based samplers including the baseline methods and our approach. On the other hand, we used 100 timesteps especially when conducting ablation studies for efficiency.

\noindent \textbf{Other details.} Our implementation is based on PyTorch~\cite{paszke2019pytorch}, and experiments were performed on twin NVIDIA A100 GPUs. Code is available at \url{https://github.com/soobin-um/sg-minority}.

\section{Additional Analyses and Discussions}
\label{sec:ablations}

\subsection{Problem formulation: a mathematical version}
We present a refined statement herein of our focused problem of \emph{generating high-quality low-likelihood instances}. Let us consider a data distribution characterized by the density function $q({\bs x}_0)$. We assume that $q$ is supported on the data manifold ${\cal M}$ that contain high-quality data instances. We further assume the continuity of $p_{\text{data}}$ across its support, under which samples with high (low) density correspond to high (low) likelihood, and vice versa. Our goal is then to generate on-manifold (i.e., high-quality) instances ${\bs x} \in {\cal M}$ that yield low density values (i.e., with low-likelihoods) under a certain threshold $\tau_{\text{th}} > 0$. More formally, it is to produce ${\bs x} \in {\cal S} \; \text{where} \; {\cal S} \coloneqq \{ {\bs x} \in {\cal M}: q({\bs x}) < \tau_{\text{th}} \}$.

\subsection{Effectiveness of the proposed metric}

\begin{figure}[!t]
\begin{center}
\includegraphics[width=1.0\columnwidth]{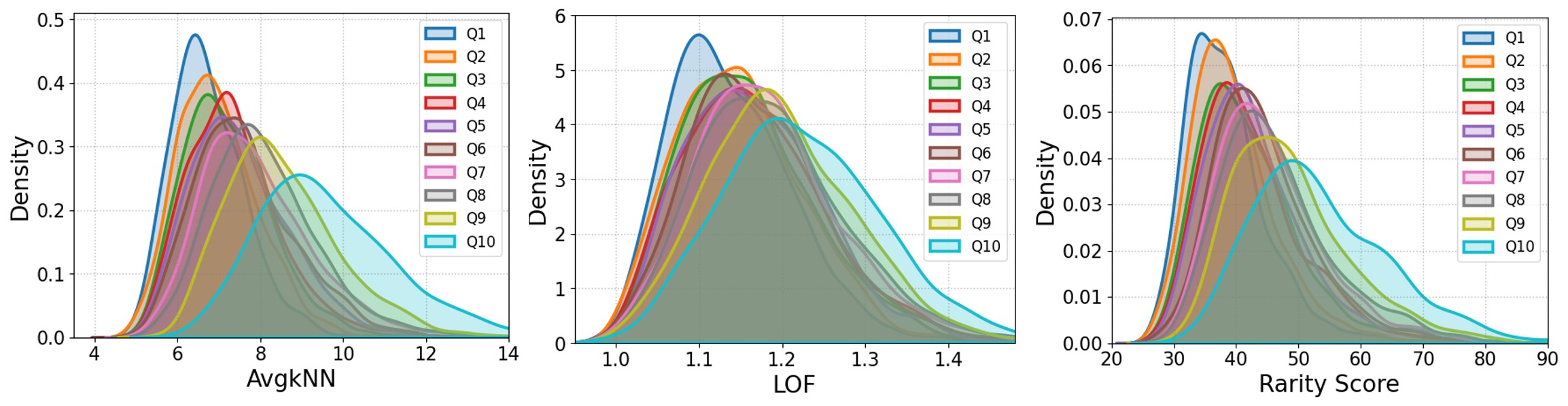}
\end{center}
\caption{Correlations of the proposed metric with existing low-likelihood measures: Average k-Nearest Neighbor (AvgkNN), Local Outlier Factor (LOF)~\cite{breunig2000lof}, and Rarity Score~\cite{han2022rarity}. For 10K CelebA generated samples by ADM~\cite{dhariwal2021diffusion}, we evaluated their uniqueness with ours and the three existing metrics. Our metric values were calculated during inference using~\cref{eq:mse}. Q\texttt{n} denotes the distribution of generated samples that yield the smallest 10\texttt{n}\% scores of~\cref{eq:mse}. For instance, Q1 in the left plot indicates the AvgkNN density of generated instances with the lowest $10\%$ of the proposed score.}
\label{fig:mse_correlations}
\end{figure}

We argued in the manuscript that our proposed metric (\ie Eq. (7)) is powerful for evaluating the uniqueness of features within intermediate latent instances ${\bs x}_t$. To support this, we provided both theoretical and empirical evidence, by showing the connection to the log-likelihood and offering the visualizations of generated samples sorted by our metric. As a further validation, we illustrate herein the correlations of the proposed metric and existing low-density metrics; see \cref{fig:mse_correlations} for details. Notice that our metric demonstrates positive correlations with the existing ones, providing an additional empirical validation as a minority metric applicable during inference time.

\subsection{Manifold-preserving aspect of the proposed guidance}

We argue that our guidance function is inherently robust to the off-manifold issue where generated samples do not lie on the data manifold ${\cal M}$. Note that this is contrary to a naive low-density guidance approach that employs $g({\bs x_t}, t) = \nabla_{{\bs x}_t} \log p_{\bs \theta} ({\bs x}_t)$ (which we mentioned in~\cref{subsec:M1}). To show this, we borrow the settings considered in~\cite{chung2022improving, chung2022diffusion} and invoke a manifold-based interpretation of diffusion models developed therein.

\begin{wrapfigure}{r}{0.4\columnwidth}
\vspace{-0.8cm}
\begin{center}
\includegraphics[width=0.4\columnwidth]{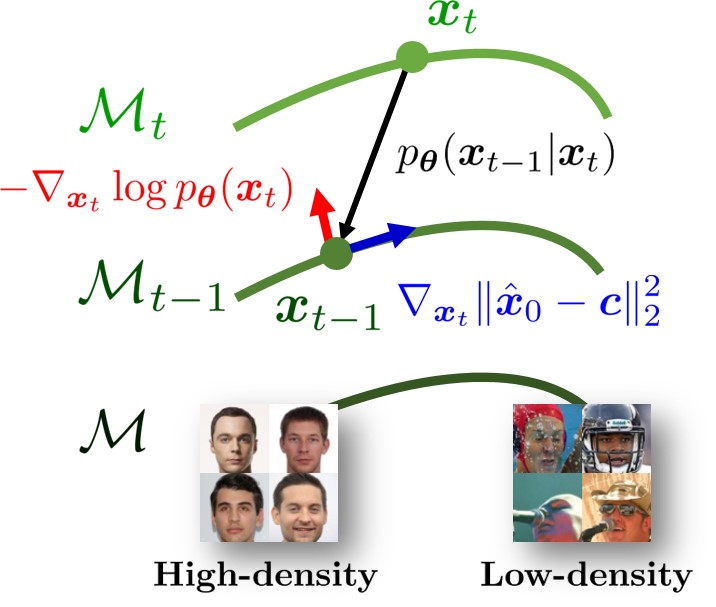}
\end{center}
\vspace{-0.3cm}
\caption{Conceptual illustration of the manifold-preserving property of our guidance}
\label{fig:mcg}
\vspace{-0.7cm}
\end{wrapfigure}

Let us consider a (clean) data manifold ${\cal M}$ constructed by a given dataset ${\bs x}_0 \sim q({\bs x}_0)$. We further consider a set of noisy manifolds $ \{ {\cal M}_t \}_{t=1}^T$ defined by the perturbed intermediate instances ${\bs x}_t \sim q_{\alpha_t}({\bs x}_t)$. As in~\cite{chung2022improving, chung2022diffusion}, we assume that the clean data manifold ${\cal M}$ is low-dimensional (compared to the ambient space) and locally linear. The forward and reverse processes of diffusion models can then be interpreted as transitions between adjacent manifolds~\cite{chung2022improving, chung2022diffusion}. For instance, the reverse process from timestep $t-1$ to $t$ can be understood as a jump from a point on ${\cal M}_{t-1}$ to another point on ${\cal M}_{t}$ (e.g., the black arrow in~\cref{fig:mcg}).
\setlength{\columnsep}{7.0pt}%
In this context, the naive low-density guidance $-\nabla_{{\bs x}_t}\log p_{\bs \theta} ({\bs x}_t)$ points perpendicular to the clean data manifold ${\cal M}$~\cite{chung2022improving, chung2022diffusion}. This potentially causes ${\bs x}_{t-1}$ to deviate from the noisy manifold ${\cal M}_{t-1}$ (e.g., the red arrow in~\cref{fig:mcg}), which could deteriorate the subsequent reverse transitions to produce the out-of-manifold generated samples. On the other hand, our guidance produces a \emph{tangent} direction to ${\cal M}_{t-1}$, thereby ensuring that instances remain on ${\cal M}_{t-1}$ (given a properly-chosen step size $w_t$). This is because our guidance term, expressible as $\nabla_{{\bs x}_t}\| \hat{{\bs x}}_0-{\bs c} \|_2^2$ (where ${\bs c}$ is a constant vector), is actually an instantiation of the manifold-constrained gradient $\nabla_{{\bs x}_t}\| {\cal A} ( \hat{{\bs x}}_0 ) - {\bs y} \|_2^2$, which has been proven to be tangent to ${\cal M}_{t-1}$ in~\cite{chung2022improving, chung2022diffusion}. Here, ${\cal A}(\cdot)$ is an arbitrary forward operator (of an inverse problem), and ${\bs y}$ is a given measurement vector. See~\cref{fig:mcg} for an illustration of the concept.

\subsection{Further ablation studies}

\begin{figure}[!t]
\begin{subfigure}[h]{0.31\textwidth}
    \vspace{0.21cm}
    \center
    \includegraphics[width=1.0\columnwidth]{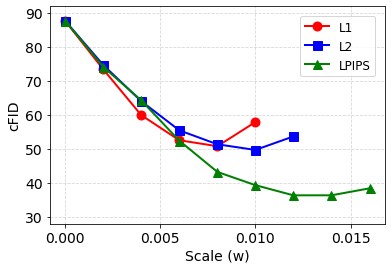}
    \caption{Impact of distance $d(\cdot, \cdot)$}
    \label{fig:ablation_d}
\end{subfigure}
\begin{subfigure}[h]{0.33\textwidth}
    \center
    \includegraphics[width=1.0\columnwidth]{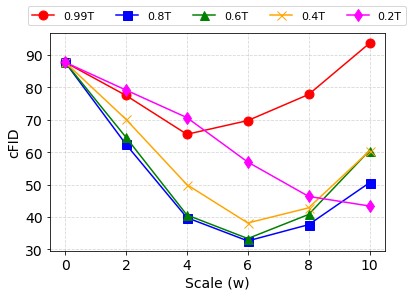}
    \caption{Influence of the timestep $s$}
    \label{fig:impact_s}
\end{subfigure}
\begin{subfigure}[h]{0.31\textwidth}
    \vspace{0.21cm}
    \center
    \includegraphics[width=1.0\columnwidth]{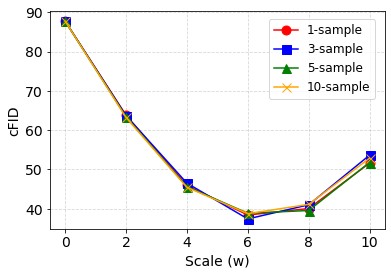}
    \caption{Effect of MC samples}
    \label{fig:impact_mc_samples}
\end{subfigure}
\caption{\textbf{(a)} Influence of the distance metric $d(\cdot, \cdot)$ in our metric in~\cref{eq:mse}. For consistent magnitudes of guidance terms over distance metrics, we normalized gradients to have unit $l_{\infty}$ norm. The use of LPIPS as $d(\cdot, \cdot)$ yields the best performance of our guided-sampler. \textbf{(b)} Impact of the perturbation timestep $s$ in~\cref{eq:mse}. $T$ indicates the total number of timesteps that the pretrained diffusion model is configured with (\ie, $T=1000$). Employing moderate levels of perturbation strength (such as $s = 0.8T$) leads to favorable performances. \textbf{(c)} Ablation on the number of samples used for the expectation in~\eqref{eq:mse}. \texttt{n-sample} indicates that $n$ random samples are drawn from $q_{\alpha_s}( \hat{\bs x}_s | \hat{\bs x}_0 )$ and employed for the Monte-Carlo estimation of the average. The number of Monte-Carlo samples is not critical to our framework, and our sampler works reasonably well with just a single drawing of instance from $q_{\alpha_s}( \hat{\bs x}_s | \hat{\bs x}_0 )$.}
\end{figure}

\noindent \textbf{Distance metric.}~\cref{fig:ablation_d} exhibits an ablation study that investigates various discrepancy metrics for $d(\cdot, \cdot)$ in our minority metric (\ie,~\cref{eq:mse}). Notice that while pixel-level distances offer significant gain when compared to the baseline ancestral sampling (\ie, the one with $w = 0$ in the figure), the use of perceptual distances like LPIPS~\cite{zhang2018perceptual} is more beneficial. This corroborates with the previous observation made in~\cite{um2023don} on the supremacy of LPIPS in capturing low-likelihood minority features.

\noindent \textbf{Perturbation timestep.}~\cref{fig:impact_s} visualizes the influence of adjusting $s$ in~\cref{eq:mse} on the performance of our approach. Notice that the use of moderate strengths of noise perturbation (\eg, $s = 0.8T$) is important for yielding good performances. We conjecture that this is due to the performance degradation of the proposed metric for differentiating low-likelihood minority features (from high-likelihood ones) when using too strong (or weak) noise perturbation $s$. Specifically for too high values of $s$ (\eg, $s = 0.99T$), $\hat{{\bs x}}_s$ could rarely preserve information on $\hat{\bs x}_0$ due to the strong injected noise. This induces significant reconstruction loss between $\hat{\bs x}_0$ and $\doublehat{{\bs x}}_0$ regardless of whether $\hat{\bs x}_0$'s features are low-likelihood or not, thereby leading to the performance deterioration of distinguishing low-likelihood features. On the other hand, when the perturbation is too weak, then almost all information of $\hat{\bs x}_0$ could remain in $\hat{{\bs x}}_s$. This enables Tweedie's formula to offer high-fidelity reconstructions $\doublehat{{\bs x}}_0$ both for high and low likelihood features within $\hat{{\bs x}}_0$, which often yields small reconstructions losses for both cases and therefore leads to degraded metric for low-likelihood features.

\noindent \textbf{Number of samples for expectation.}~\cref{fig:impact_mc_samples} illustrates the impact of the number of number of samples employed for estimating the average in our metric in~\cref{eq:mse}. We see that the performance of our guidance is not that sensitive to the number of samples for the Monte-Carlo estimation. In fact, as described in~\cref{sec:imple}, all our main results (\eg, in~\cref{table:metrics}) were derived with the use of a single Monte-Carlo sample, further demonstrating its efficient aspect that offers significant gain without heavy computations.

\begin{figure}[!t]
\begin{subfigure}[h]{0.48\textwidth}
    \centering
    \vspace{0.5cm}
    \fontsize{8}{8}\selectfont
        {
        \scalebox{1.0}{
    \begin{booktabs}{colspec = {Q[l, 0.5cm]Q[c, 0.65cm]Q[c, 0.65cm]Q[c, 0.63cm]Q[c, 0.63cm]}}
        \toprule
        $t_{\text{mid}}$ & cFID & sFID & Prec & Rec \\
        \midrule
        $0.0$ & 41.78 & 17.61 & 0.74 & 0.56 \\
        $0.1T$ & 44.24 & 17.08 & 0.79 & 0.56 \\
        $0.2T$ & 45.46 & 16.79 & 0.83 & 0.52 \\
        $0.3T$ & 46.46 & 17.14 & 0.85 & 0.50 \\
        \bottomrule
    \end{booktabs}
    }
    }
    \vspace{0.52cm}
    \caption{Impact of switch-off}
    \label{tab:so_t}
\end{subfigure}
\begin{subfigure}[h]{0.48\textwidth}
    \center
    \includegraphics[width=1.0\columnwidth]{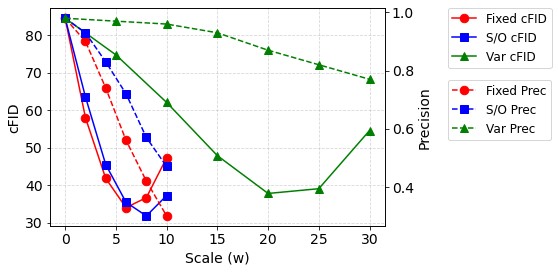}
    \caption{Ablation on the time-schedules}
    \label{fig:ablation_ts}
\end{subfigure}
\caption{\textbf{(a)} Effect of incorporating early stopping in our guidance approach. $t_{\text{mid}}$ is a threshold parameter that determines the timestep from which our low-likelihood guidance term is deactivated afterwards. $T$ indicates the total number of timesteps with which the pretrained diffusion model is configured (\eg, $T=1000$ for CelebA). The sample quality due to our guided sampler can be significantly improved by early stopping (or gradually decreasing) the strength of the guidance term. \textbf{(b)} Ablation on time-scheduling strategies. \texttt{Fixed} indicates our sampler with a fixed scale over time (\ie, $w_t = w$). \texttt{S/O} denotes the case employing sudden switch-off: $w_t = w \cdot \mathbbm{1}\{t \ge t_{\text{mid}} \}$. We used $t_{\text{mid}} = 0.2T$ for the results exhibited herein. \texttt{Var} is the variance-based time-scheduling: $w_t = w \cdot {\bs \Sigma}_{\bs \theta} ( {\bs x}_t, t )$. Solid lines indicate cFID~\cite{parmar2022aliased} performances while dotted lines are performance values of Improved Precision~\cite{kynkaanniemi2019improved}. The most balanced schedule is \texttt{Var}, \ie, the ones that leverage the noise statistics of the pretrained model.}
\end{figure}

\noindent \textbf{Time-scheduling strategies.}~\cref{tab:so_t} ablates the threshold parameter $t_{\text{mid}}$ of the sudden switch-off scheduling. We see great advantage in early stopping our guidance term, yielding significant gain in sample quality while incurring marginal degradation in diversity (\eg, when comparing the performances of $t_{\text{mid}} = 0.0$ and $t_{\text{mid}} = 0.1T$). This offers an empirical evidence that validates our motivation of developing gradually-decreasing time schedules in~\cref{subsec:method_3}.

\cref{fig:ablation_ts} provides an investigation on our proposed time-schedules of our guidance term. While all schedules offer significant improvements in sample quality over the case with fixed scales (reflected in higher values of precision), we observe that the schedule built upon the reverse diffusion process (\ie, \texttt{Var}) yields favorable trade-offs between sample quality and diversity when compared to unprincipled ones (like \texttt{S/O}).

\begin{wraptable}{r}{0.5\textwidth}
\vspace{-1.1cm}
\caption{Impact of the intermittent rate $n$ on performance. ``Time'' indicates inference time measured in sec/sample.}
\centering
\fontsize{8}{8}\selectfont
    {
    \scalebox{0.9}{
\begin{booktabs}{colspec = {Q[l, 0.2cm]Q[c, 0.65cm]Q[c, 0.65cm]Q[c, 0.6cm]Q[c, 0.6cm] | Q[c, 0.6cm]}}
    \toprule
    $n$ & cFID & sFID & Prec & Rec & Time \\
    \midrule
    1 & 39.09 & 43.68 & 0.85 & 0.52 & 1.75 \\
    2 & 40.97 & 43.92 & 0.82 & 0.52 & 1.09  \\
    5 & 40.29 & 43.01 & 0.83 & 0.51 & 0.67 \\
    10 & 44.16 & 44.36 & 0.84 & 0.50 & 0.55 \\
    20 & 47.51 & 46.95 & 0.80 & 0.51 & 0.47 \\
    \bottomrule
\end{booktabs}
}
}
\vspace{-0.8cm}
\label{tab:inter_rate}
\end{wraptable}

\noindent \textbf{Intermittent rate.}~\cref{tab:inter_rate} demonstrates the impact of the intermittent rate $n$ on our performance metrics. Observe that the use of the intermittent technique significantly improves resource efficiency while maintaining performance benefits of ours upto $n=5$. We highlight that the adoption of the intermittent technique enables our approach to achieve competitive computational costs compared to existing techniques~\cite{sehwag2022generating, um2023don}. See~\cref{tab:comparison_complexities} for a detailed complexity comparison.

\subsection{Controllability of the proposed approach}

\setlength{\columnsep}{7.0pt}%
\begin{wrapfigure}{r}{0.5\columnwidth}
\vspace{-0.8cm}
\begin{center}
\includegraphics[width=0.5\columnwidth]{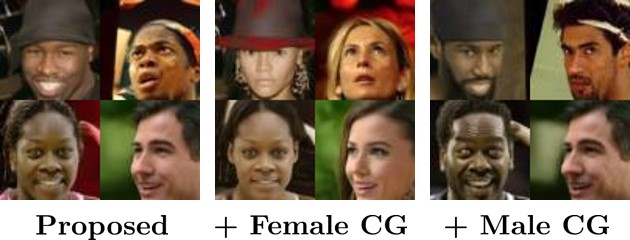}
\end{center}
\vspace{-0.3cm}
\caption{Controllable nature of the proposed approach. Generated samples by our method (left), ours with classifier guidance (CG) toward ``female'' (middle) and ``male'' (right) are exhibited.}
\label{fig:controllability}
\vspace{-0.7cm}
\end{wrapfigure}
One may concern that the proposed guidance approach could lose some controllability (e.g., over semantics) potentially offered by previous classifier-based methods~\cite{sehwag2022generating, um2023don}. However, we contend that the controllability of our approach does not fall behind the prior works. Specifically given an external classifier capable of recognizing desired semantics (e.g., a gender predictor), our method enables semantically-controlled low-likelihood generation by integrating classifier guidance (CG) into our sampler; see~\cref{fig:controllability} for instance on CelebA.

\subsection{Computational complexities}

\cref{tab:comparison_complexities} presents a comparison of computational burdens between our approach and focused baselines using the CelebA dataset. Thanks to the intermittent technique, we achieve competitive inference times comparable to existing samplers, while maintaining superior performance in generating minority samples compared to the baselines (see~\cref{table:metrics} for detailed performance values). Notice that significant additional resources (\eg, dozens of \textbf{hours}) are required when incorporating the baseline minority samplers like Sehwag \etal~\cite{sehwag2022generating}; details on the derivation of these loads are provided below. We emphasize that the extra burdens could be exacerbated especially for more complicated benchmarks with larger scales. For instance, as mentioned in the introduction, the additional time costs of the both methods~\cite{sehwag2022generating, um2023don} for the ImageNet-64 results were more than \textbf{40} V100-days~\cite{dhariwal2021diffusion}. In contrast, our sampler incurs no such preparation overhead while offering comparable inference times, making it a more practical solution.

\begin{wraptable}{r}{0.5\textwidth}
    \vspace{-1.1cm}
    \caption{Complexity comparison with existing samplers. ``Infer'' indicates inference time measured in sec/sample. ``Extra'' indicates \textbf{hours} required for constructing external classifiers. All quantities were measured using a single NVIDIA A100 GPU. The use of the intermittent technique enables our approach to achieve competitive inference cost compared to existing minority samplers, all the while avoiding the introduction of any supplementary expenses.}
    \label{tab:comparison_complexities}
    \centering
    \fontsize{9}{9}\selectfont
        {
        \scalebox{1.0}{
    \begin{booktabs}{colspec = {Q[l, 3.3cm]Q[c, 0.63cm]Q[c, 0.75cm]}}
        \toprule
          Method  & Infer & Extra \\
        \midrule
        ADM~\cite{dhariwal2021diffusion}  & 0.43s & -- \\
        Sehwag \etal~\cite{sehwag2022generating}  & 0.71s & 51.05h \\
        Um and Ye\cite{um2023don}  & 0.57s & 2.84h \\
        Ours (+ intermittent)  & 0.67s & -- \\
        \bottomrule
    \end{booktabs}
    }
    }
    \vspace{-0.6cm}
\end{wraptable}

We leave details herein on the evaluations of the additional loads of the baselines. For~\cite{sehwag2022generating} in~\cref{tab:comparison_complexities}, we spent 10.61 hours for training the classifier used for pushing instances to low-likelihood regions.~\cite{sehwag2022generating} also employs a real-fake discriminator for improving sample quality, and its training requires significant number of fake samples generated by a given pretrained diffusion model. This leaded to additional 29.38 hours for the generation of fake samples and 11.06 hours for the subsequent discriminator training, thereby spending total 51.04 hours for~\cite{sehwag2022generating}. For~\cite{um2023don} in~\cref{tab:comparison_complexities}, we first spent 0.92 hours to construct the training dataset for the minority classifier, which includes the labeling of the given CelebA samples with minority score. Subsequently for training the classifier, we used 1.92 hours, thereby yielding total 2.84 hours. For the ImageNet-64 experiments, both methods~\cite{sehwag2022generating, um2023don} employed the classifier developed in~\cite{dhariwal2021diffusion}, for which the authors in~\cite{dhariwal2021diffusion} invested 40 V-100 days in training; see Tab. 10 and 12 in~\cite{dhariwal2021diffusion} for details.

\begin{figure}[!t]
    \begin{subfigure}[h]{0.321\textwidth}
    \centering
    \includegraphics[width=1.0\columnwidth]{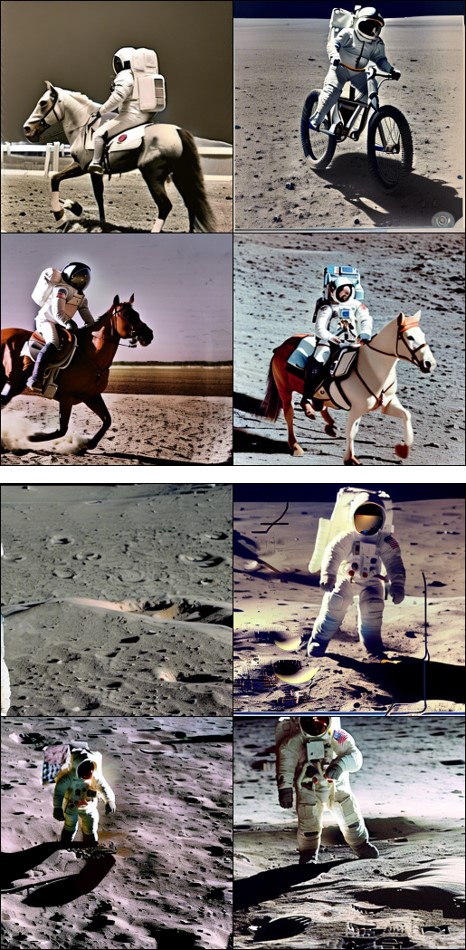}
    \caption{DDIM~\cite{song2020denoising}}
    \end{subfigure}
    \hfill
    \begin{subfigure}[h]{0.321\textwidth}
    \centering
    \includegraphics[width=1.0\columnwidth]{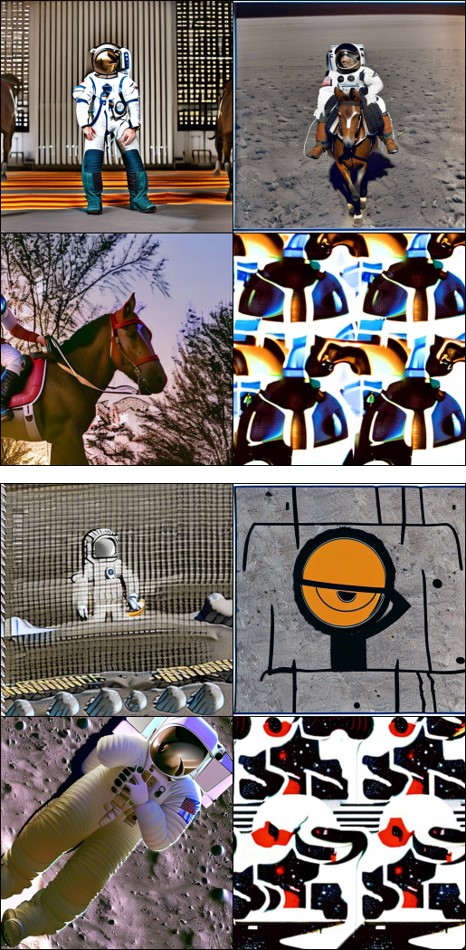}
    \caption{CADS~\cite{sadat2023cads}}
    \end{subfigure}
    \hfill
    \begin{subfigure}[h]{0.321\textwidth}
    \centering
    \includegraphics[width=1.0\columnwidth]{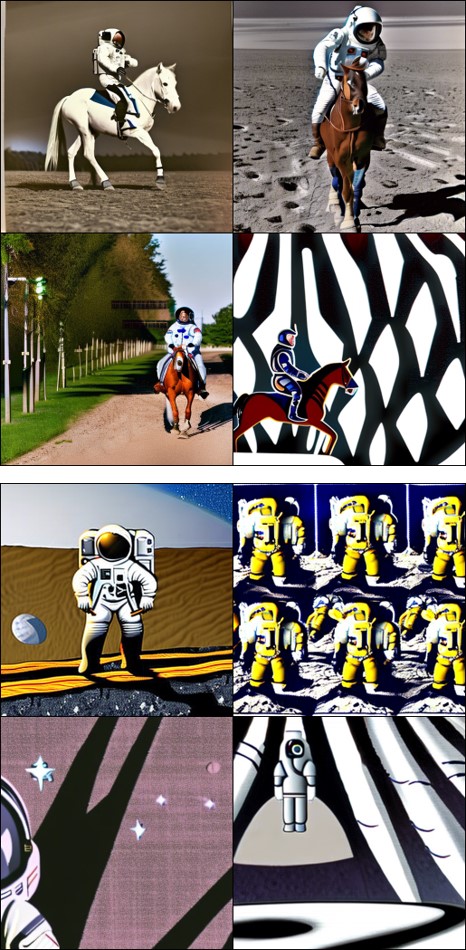}
    \caption{Ours}
    \end{subfigure}
\caption{Generated samples in the context of T2I generation. Two distinct prompts are considered: (i) \emph{``A professional photograph of an astronaut riding a horse''} (top row); (ii) \emph{``An astronaut on the Moon''} (bottom row). For each row, we share the same random seed across all three methods. The generated samples by our approach are more prone to contain low-likelihood minority attributes (\eg, with revealing unique artistic features~\cite{serra2019input, arvinte2023investigating}).}
\label{fig:t2i_samples}
\end{figure}

\section{Further Applications}
\label{sec:further_applications}

\subsection{Text-to-image generation}

We demonstrate the practical significance of our approach herein by investigating the application on text-to-image (T2I) generation -- a challenging-yet-important task that draws substantial attention these days. Specifically, our goal is to create low-likelihood minority images w.r.t. given prompts, which are rarely produced via standard sampling techniques. To do so, we incorporate our guidance term into the standard sampling process of Stable Diffusion~\cite{rombach2022high} (v2.1).

\begin{wrapfigure}{r}{0.5\textwidth}
\centering
\includegraphics[width=0.5\textwidth]{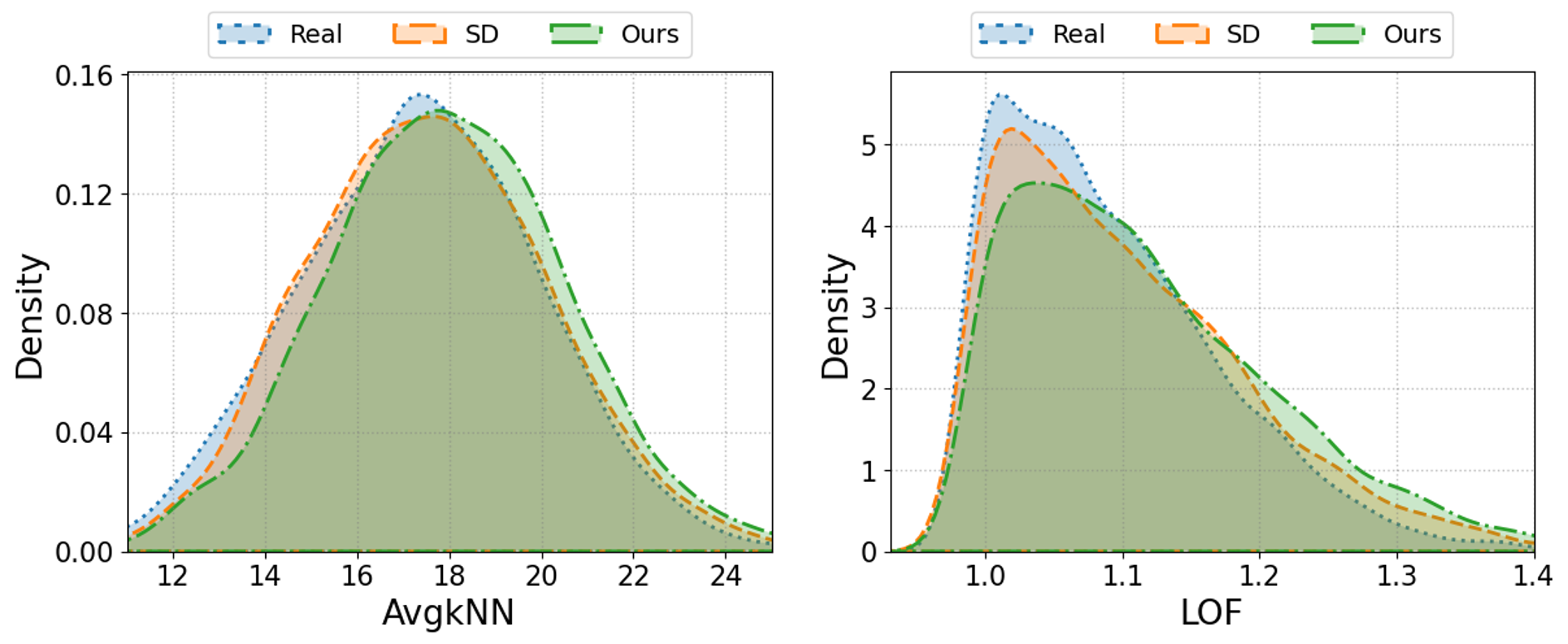}
\caption{Comparison of neighborhood density on the T2I generation task. ``Real'' indicates the validation set of MS-COCO~\cite{lin2014microsoft}, and ``SD'' denotes Stable Diffusion~\cite{rombach2022high}. Our approach produces a greater number of minority samples compared to the standard sampler of Stable Diffusion.}
\label{fig:t2i_densities}
\vspace{-0.5cm}
\end{wrapfigure}

\cref{fig:t2i_samples} exhibits generated samples by three distinct methods. For the generations, we used two distinct prompts: (i) \emph{``A professional photograph of an astronaut riding a horse''}; (ii) \emph{``An astronaut on the Moon''}. Observe that the generated samples due to our approach are more likely to contain unique minority attributes that are often characterized by exquisite and complex visual aspects~\cite{serra2019input, arvinte2023investigating}. The observation is corroborated by the results in~\cref{fig:t2i_densities} where we see that our approach produces more low-likelihood instances (having higher values of AvgkNN and LOF) compared to the baseline sampler of Stable Diffusion. This demonstrates the effectiveness of our approach even in challenging applications such as T2I, thereby enriching the landscape of image generation.

\begin{figure}[!t]
\begin{subfigure}[h]{0.321\textwidth}
    \centering
    \includegraphics[width=1.0\columnwidth]{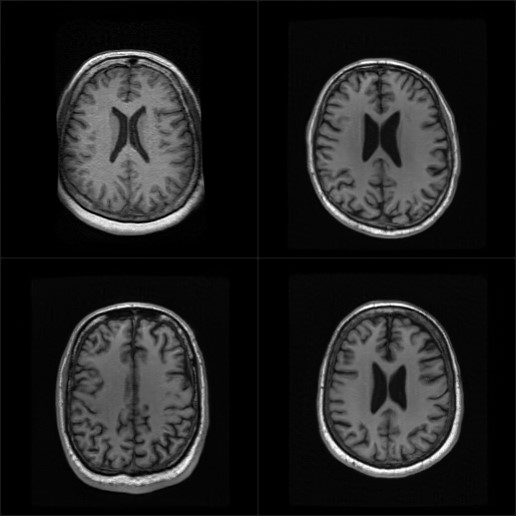}
    \caption{StyleGAN2-ADA~\cite{Karras2020ada}}
    \end{subfigure}
    \hfill
    \begin{subfigure}[h]{0.321\textwidth}
    \centering
    \includegraphics[width=1.0\columnwidth]{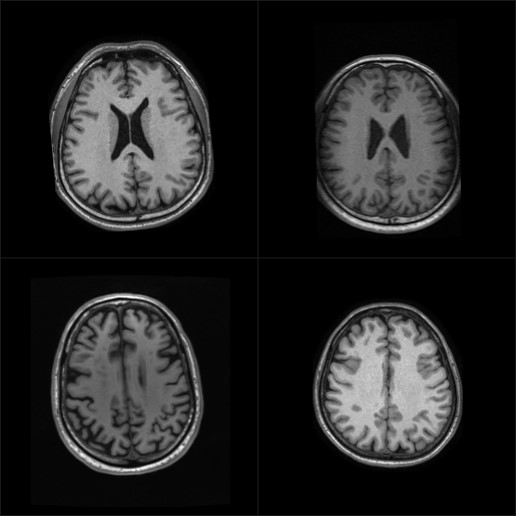}
    \caption{ADM~\cite{dhariwal2021diffusion}}
    \end{subfigure}
    \hfill
    \begin{subfigure}[h]{0.321\textwidth}
    \centering
    \includegraphics[width=1.0\columnwidth]{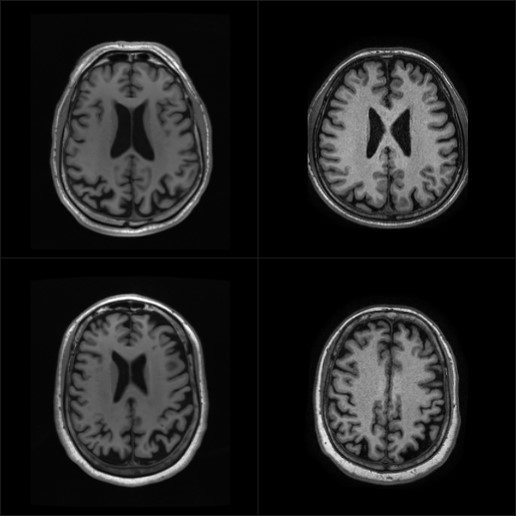}
    \caption{Ours}
    \end{subfigure}
\caption{Comparison of generated samples on our focused brain MRI dataset. The same random seed was employed for the diffusion-based samplers, \ie, middle and right. The samples generated by our approach exhibit more pronounced brain atrophy in visual aspects compared to the baseline results.}
\label{fig:mri_samples}
\end{figure}

\subsection{Medical imaging}

\begin{wraptable}{r}{0.53\textwidth}
    \vspace{-1.1cm}
    \caption{Comparison of sample quality and diversity on our brain MRI dataset. For baseline real data, we employ the most unique samples that yield the highest AvgkNN values. Our framework demonstrates improvement over the baselines without relying on external elements.}
    \label{table:metrics_mri}
    \centering
    \fontsize{8}{8}\selectfont
    {
    \scalebox{1.0}{
    \begin{tabular}{lcccc}
        \toprule[0.1em]
          \multicolumn{1}{l}{Method}  & \multicolumn{1}{c}{cFID} & \multicolumn{1}{c}{sFID} & \multicolumn{1}{c}{Prec} & \multicolumn{1}{c}{Rec} \\
          \\
        \multicolumn{5}{l}{\textbf{Brain MRI 256$\times$256}}\\
        \toprule[0.1em]
        \multicolumn{1}{l}{ADM~\cite{dhariwal2021diffusion}} & \underline{22.51} & \underline{16.02} & \underline{0.77} & \underline{0.55} \\
        \multicolumn{1}{l}{StyleGAN2-ADA~\cite{Karras2020ada}} & 23.07 & 23.67 & 0.52 & 0.49 \\
        \midrule[0.001em]
        \multicolumn{1}{l}{Ours} & \textbf{22.05} & \textbf{15.91} & \textbf{0.78} & \textbf{0.56}
    \end{tabular}
    }
    }
    \vspace{-0.6cm}
\end{wraptable}
To demonstrate a broad applicability of our approach, we push the boundary beyond natural images and explore the domain of medial imaging. Specifically, we consider an in-house (IRB-approved) brain MRI dataset containing 13,640 axial slice images, where low-likelihood instances are ones that exhibit degenerative brain disease like cerebral atrophy. The MRI images are standard 3T T1-weighted with $256^2$ resolution. Our brain MRI experiments were conducted with two baselines. The first one is ADM~\cite{dhariwal2021diffusion} with ancestral sampling~\cite{ho2020denoising}. The second baseline is StyleGAN2-ADA~\cite{Karras2020ada}, a powerful GAN-based framework that has demonstrated its effectiveness in medical imaging applications~\cite{woodland2022evaluating}. We constructed the pretrained backbone by ourselves by respecting the same architecture and the setting used for LSUN-Bedrooms in~\cite{dhariwal2021diffusion}. To obtain the baseline StyleGAN2-ADA model, we respected the settings provided in the official codebase\footnote{\url{https://github.com/NVlabs/stylegan2-ada-pytorch}} and trained the model by ourselves. As our other experiments, we evaluated sample quality and diversity by comparing generated samples with low-likelihood real data yielding the highest AvgkNN values.

\cref{fig:mri_samples} exhibits generated samples by three distinct methods on the brain MRI benchmark. Observe that our sampler are more likely to produce low-likelihood features of the dataset (\eg, containing severe brain atrophy) compared to the baseline methods. We emphasize that our capability of generating low-likelihood instances persists even on this particular type of data containing distinctive visual aspects, further demonstrating the practical significance of our method. The quantitative results in~\cref{table:metrics_mri} support this observation. Notice in the figure that our framework improves over the baselines in terms of minority-generating capability, highlighting the robustness and versatility of our approach beyond the domain of natural images.

\subsection{Image editing}

\begin{wrapfigure}{r}{0.5\textwidth}
\vspace{-0.7cm}
\centering
\includegraphics[width=0.5\textwidth]{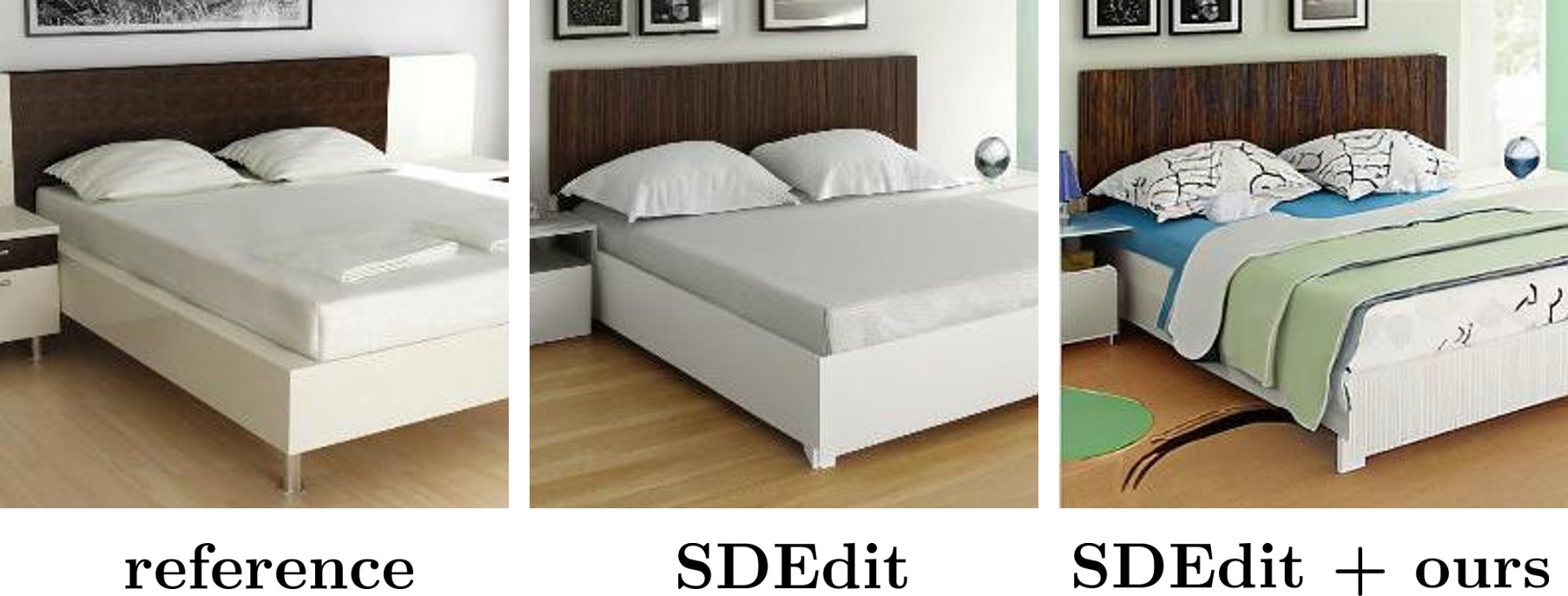}
\caption{Application in image editing. Our method introduces novel attributes into the reference image, which is difficult to achieve with the baseline editing framework.}
\label{fig:editing}
\vspace{-0.8cm}
\end{wrapfigure}
We investigate the application of our approach in image editing, one prominent application area of generative models widely employed in practice. The interest herein is to introduce distinctive elements into a target reference image, which is a key focus in industries such as creative AI~\cite{rombach2022high, han2022rarity}. To this end, we integrate our approach into the editing pipeline of SDEdit~\cite{meng2021sdedit}. Specifically, we incorporate the proposed guidance term into the reverse process of the SDEdit pipeline to yield minority features during reconstruction.

\cref{fig:editing} visualizes the application of our approach on LSUN-Bedrooms. Observe that our method introduces novel visual attributes when compared to the baseline SDEdit framework. We highlight that this demonstrates the practical importance of our work and its potential applicability across a wide variety of practical scenarios.

\begin{figure}[!t]
    \begin{subfigure}[h]{0.321\textwidth}
    \centering
    \includegraphics[width=1.0\columnwidth]{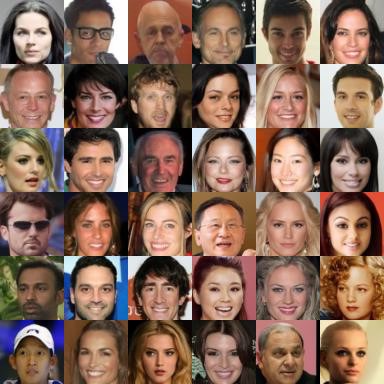}
    \caption{ADM~\cite{dhariwal2021diffusion}}
    \end{subfigure}
    \hfill
    \begin{subfigure}[h]{0.321\textwidth}
    \centering
    \includegraphics[width=1.0\columnwidth]{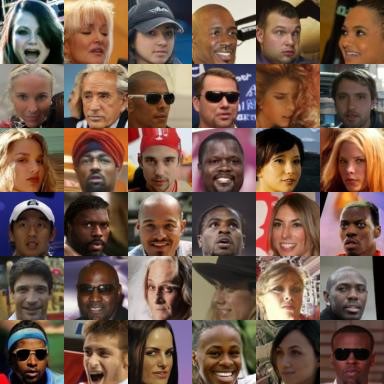}
    \caption{Um and Ye~\cite{um2023don}}
    \end{subfigure}
    \hfill
    \begin{subfigure}[h]{0.321\textwidth}
    \centering
    \includegraphics[width=1.0\columnwidth]{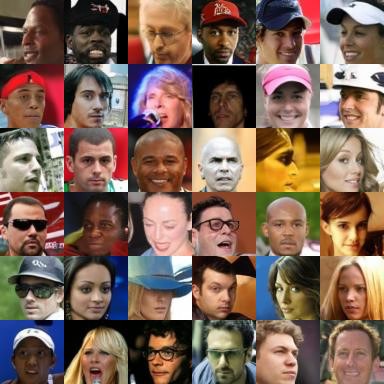}
    \caption{Ours}
    \end{subfigure}
    \vspace{-0.3cm}
\caption{Sample comparison on CelebA. We share the same random seed across all three methods.}
\label{fig:celeba_samples}
\end{figure}

\section{Additional Experimental Results}
\label{sec:additional_results}

\noindent \textbf{Generated samples on CelebA.} \cref{fig:celeba_samples} visualizes generated samples on CelebA. Notice that the generated samples by both minority samplers are more likely to contain unique features of the dataset compared to the samples from a standard sampler. However, we highlight that our performance gain stems exclusively from the pretrained model, which is in stark contrast with the other minority sampler helped by an external classifier to yield the minority-enhanced generation.

\begin{figure}[!t]
\centering
\includegraphics[width=1.0\columnwidth]{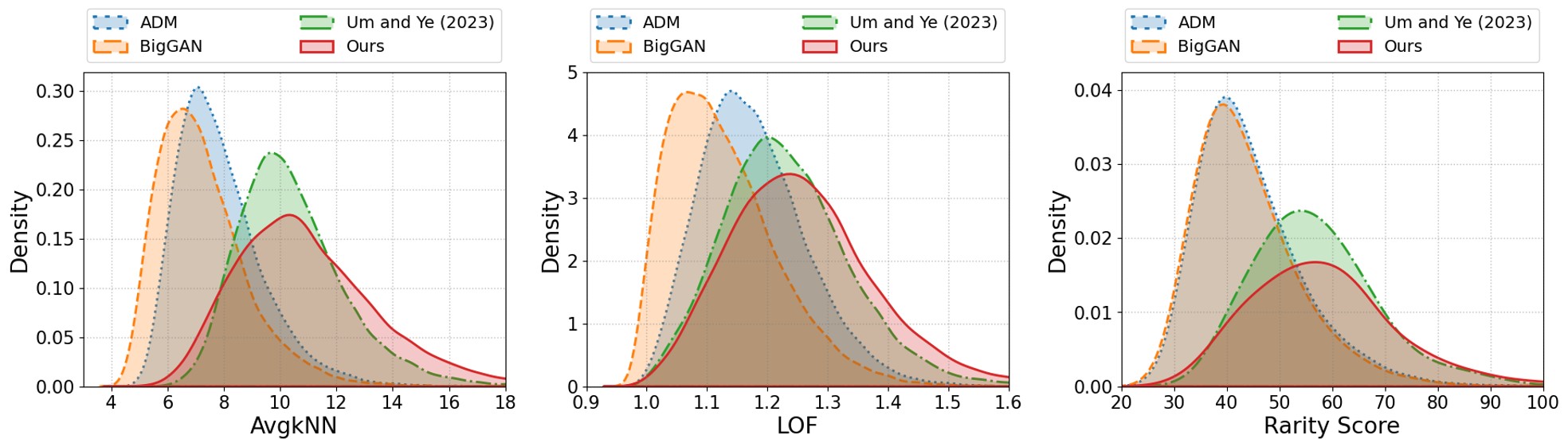}
\vspace{-0.6cm}
\caption{Comparison of neighborhood density on CelebA. “AvgkNN” refers to Average k-Nearest Neighbor, and “LOF” is Local Outlier Factor~\cite{breunig2000lof}. ``Rarity Score'' indicates a low-density metric proposed by~\cite{han2022rarity}. The higher values, the less likely samples for all three measures.}
\label{fig:celeba_densities}
\vspace{-0.3cm}
\end{figure}

\begin{figure}[!t]
\begin{center}
\includegraphics[width=1.0\columnwidth]{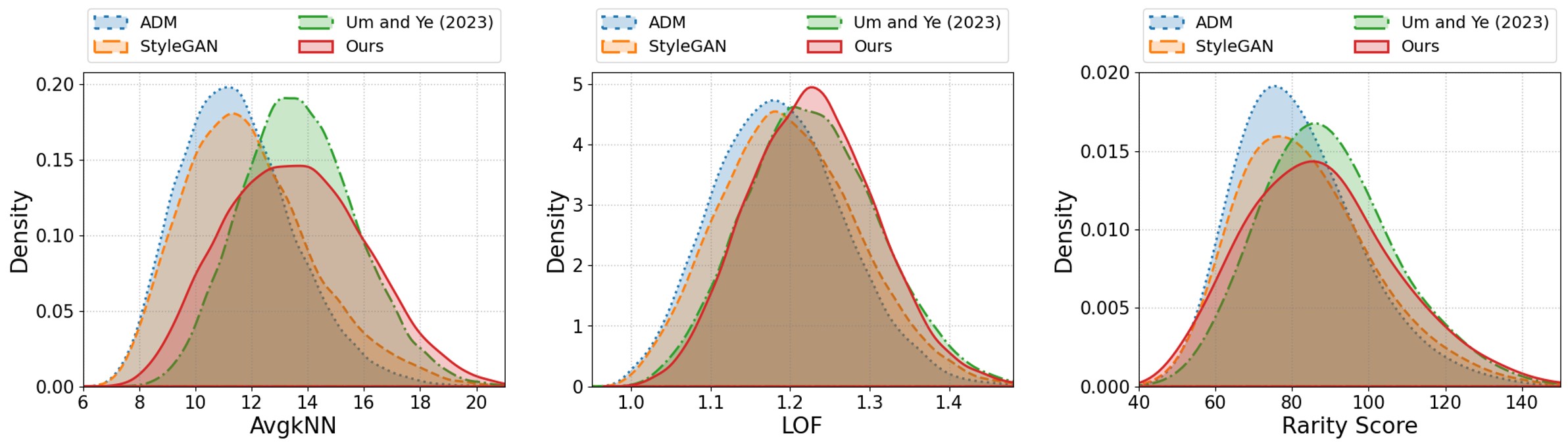}
\end{center}
\vspace{-0.6cm}
\caption{Comparison of neighborhood density on LSUN-Bedrooms. The results are the same as those in~\cref{fig:lsun_densities}.}
\label{fig:lsun_densities_apdx}
\vspace{-0.3cm}
\end{figure}

\begin{figure}[!t]
\centering
\includegraphics[width=1.0\columnwidth]{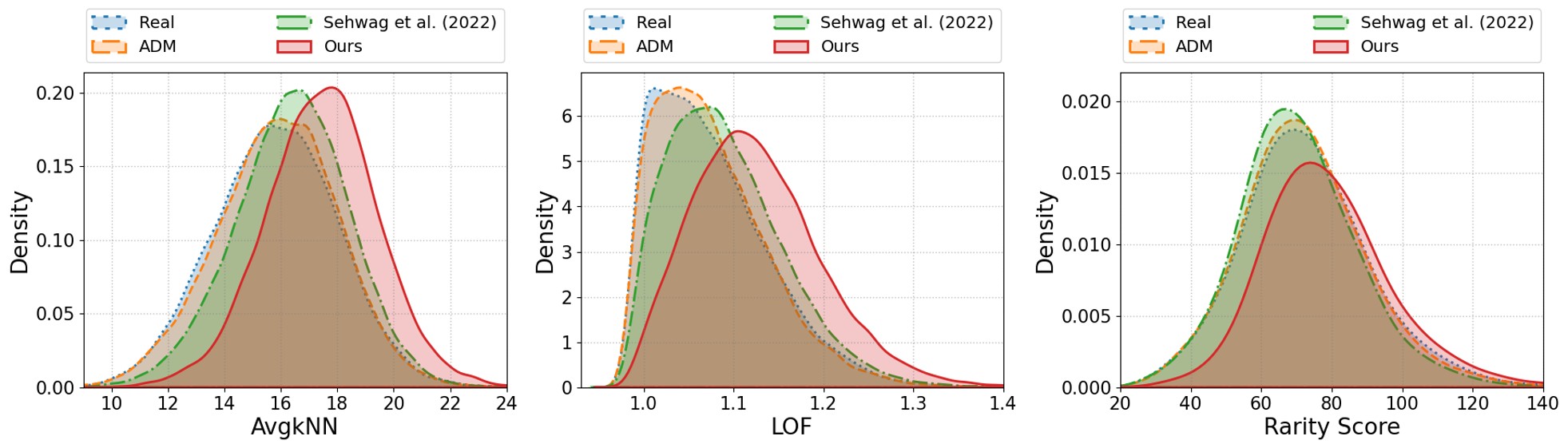}
\vspace{-0.6cm}
\caption{Comparison of neighborhood density on ImageNet-64. All the settings are the same as those in Figure~\ref{fig:celeba_densities}.}
\label{fig:imagenet64_densities}
\vspace{-0.3cm}
\end{figure}

\begin{figure}[!t]
\centering
\includegraphics[width=1.0\columnwidth]{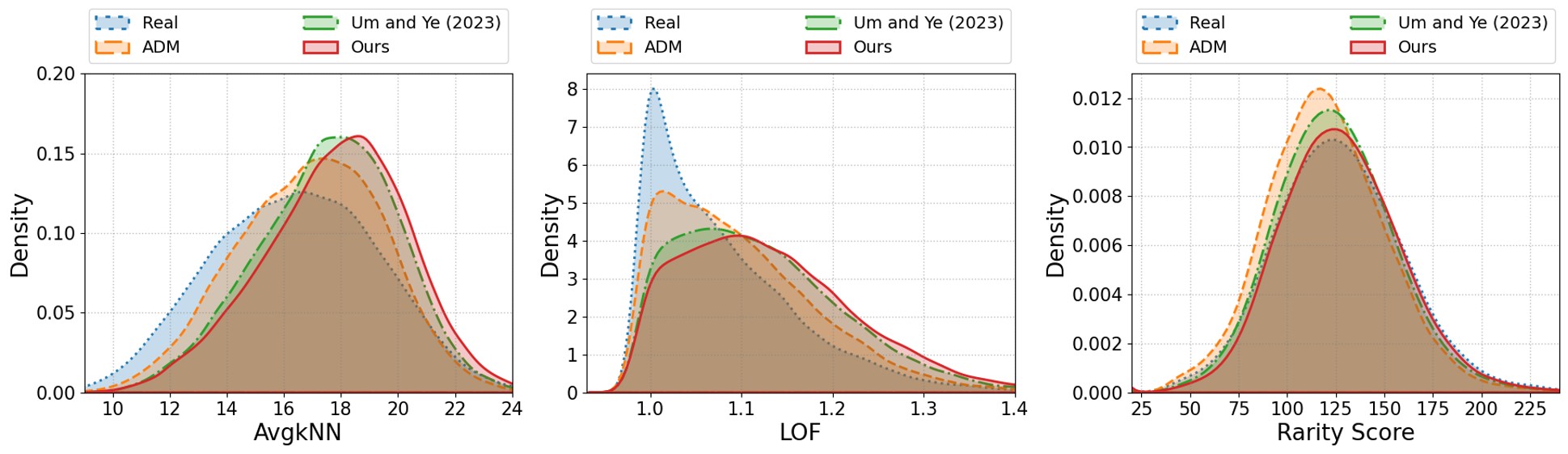}
\vspace{-0.6cm}
\caption{Comparison of neighborhood density on ImageNet-256. All the settings are the same as those in Figure~\ref{fig:celeba_densities}.}
\label{fig:imagenet256_densities}
\vspace{-0.3cm}
\end{figure}

\noindent \textbf{Density results on other datasets.} \cref{fig:celeba_densities}, \ref{fig:imagenet64_densities}, and \ref{fig:imagenet256_densities} illustrate the neighborhood density outcomes on CelebA, ImageNet-64 and 256 respectively. Note that across all three metrics evaluated in these benchmarks, our self-guided sampler consistently outperforms or achieves comparable performance to the baselines in generating low-likelihood minority instances. This highlights the generic advantages of our approach, which are not limited to a specific dataset. For completeness, we include herein the neighborhood density results of LSUN-Bedrooms, which are already reported in the manuscript; see~\cref{fig:lsun_densities_apdx} for details.

\begin{figure}[!t]
    \begin{subfigure}[h]{0.321\textwidth}
    \centering
    \includegraphics[width=1.0\columnwidth]{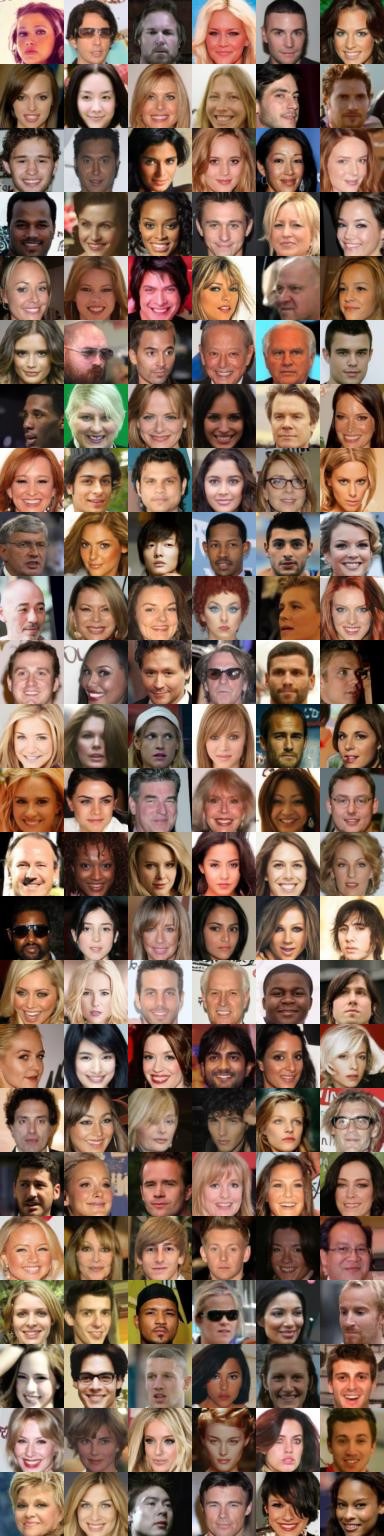}
    \caption{ADM~\cite{dhariwal2021diffusion}}
    \end{subfigure}
    \hfill
    \begin{subfigure}[h]{0.321\textwidth}
    \centering
    \includegraphics[width=1.0\columnwidth]{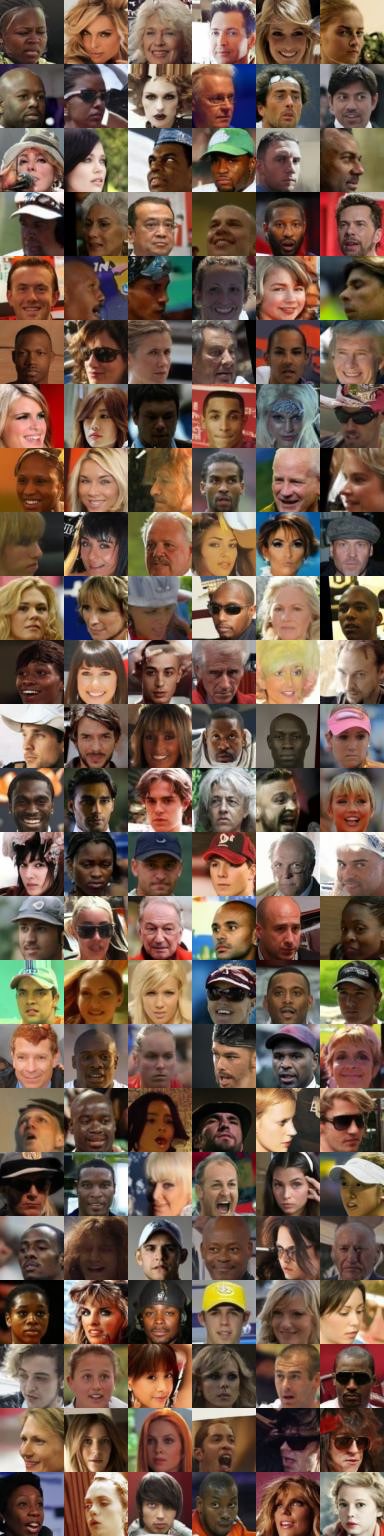}
    \caption{Um and Ye~\cite{um2023don}}
    \end{subfigure}
    \hfill
    \begin{subfigure}[h]{0.321\textwidth}
    \centering
    \includegraphics[width=1.0\columnwidth]{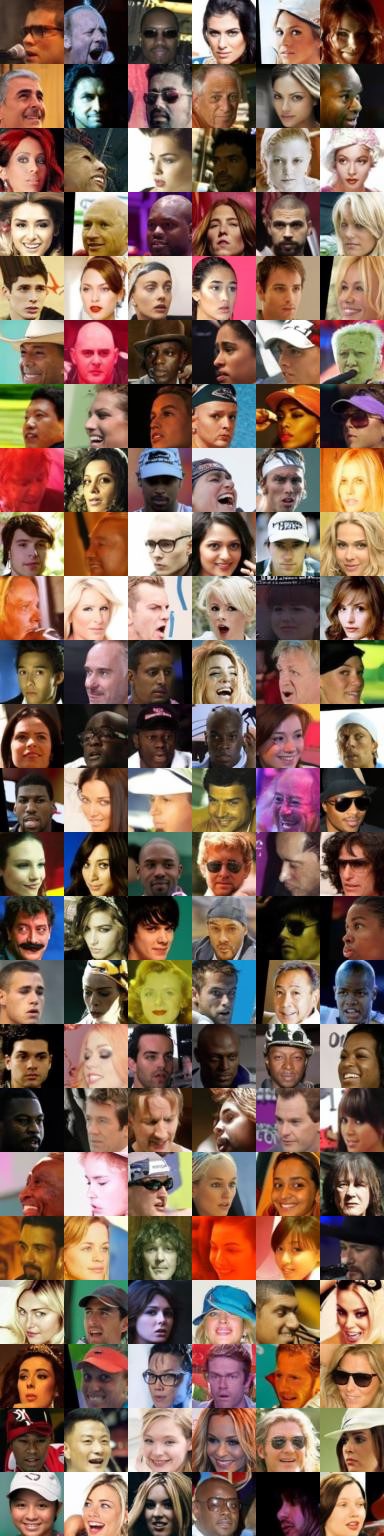}
    \caption{Ours}
    \end{subfigure}
\caption{Additional comparison of generated samples on CelebA.}
\label{fig:many_celeba_samples}
\end{figure}

\begin{figure}[!t]
    \begin{subfigure}[h]{0.321\textwidth}
    \centering
    \includegraphics[width=1.0\columnwidth]{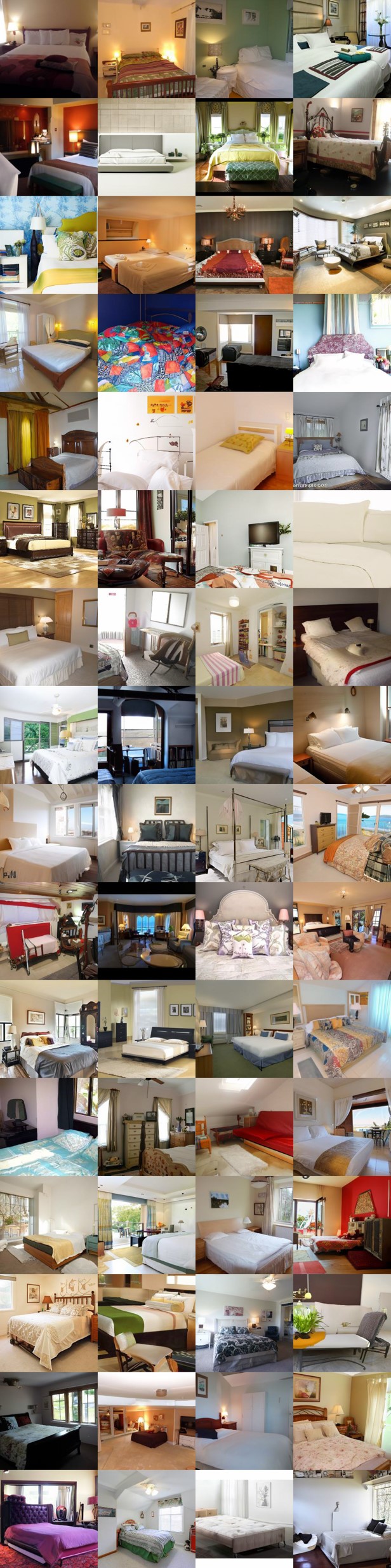}
    \caption{ADM~\cite{dhariwal2021diffusion}}
    \end{subfigure}
    \hfill
    \begin{subfigure}[h]{0.321\textwidth}
    \centering
    \includegraphics[width=1.0\columnwidth]{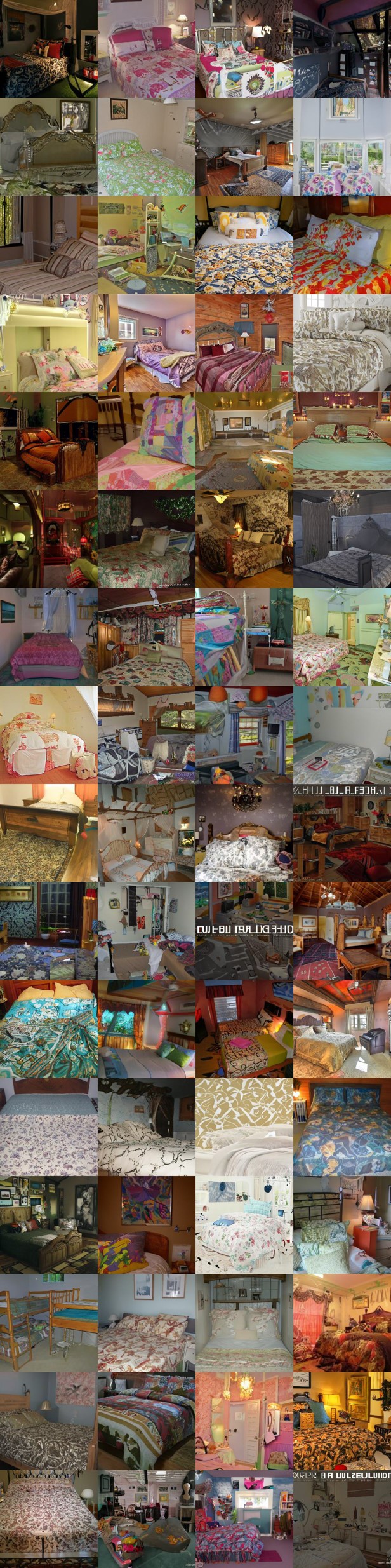}
    \caption{Um and Ye~\cite{um2023don}}
    \end{subfigure}
    \hfill
    \begin{subfigure}[h]{0.321\textwidth}
    \centering
    \includegraphics[width=1.0\columnwidth]{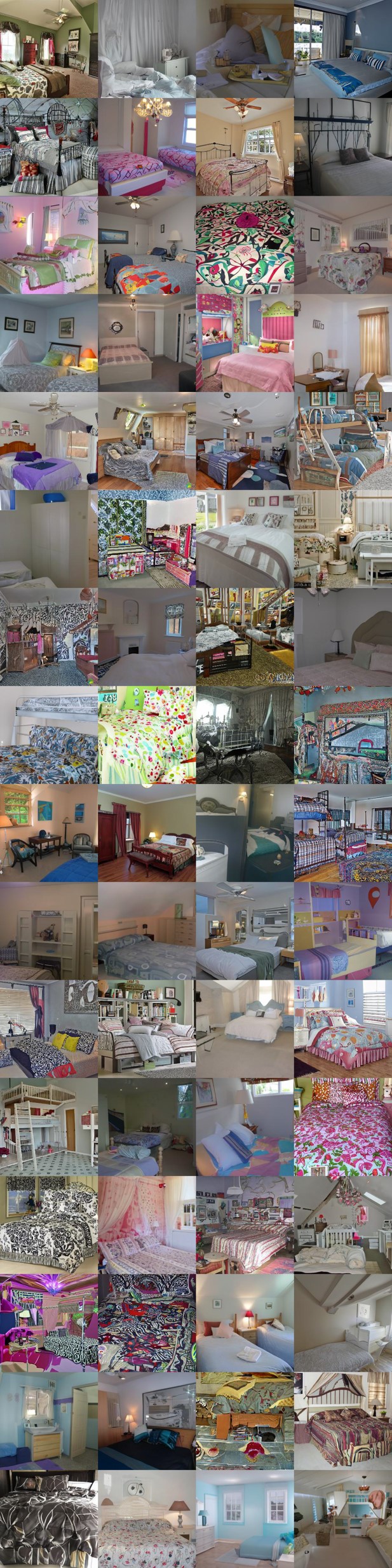}
    \caption{Ours}
    \end{subfigure}
\caption{Additional comparison of generated samples on LSUN-Bedrooms.}
\label{fig:many_lsun_samples}
\end{figure}

\begin{figure}[!t]
    \begin{subfigure}[h]{0.321\textwidth}
    \centering
    \includegraphics[width=1.0\columnwidth]{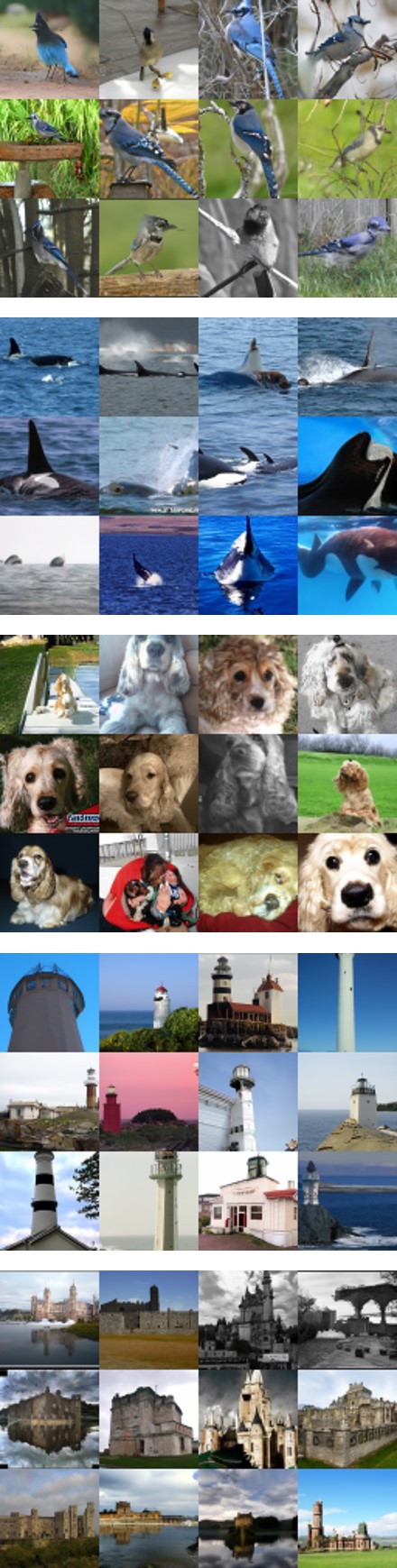}
    \caption{ADM~\cite{dhariwal2021diffusion}}
    \end{subfigure}
    \hfill
    \begin{subfigure}[h]{0.321\textwidth}
    \centering
    \includegraphics[width=1.0\columnwidth]{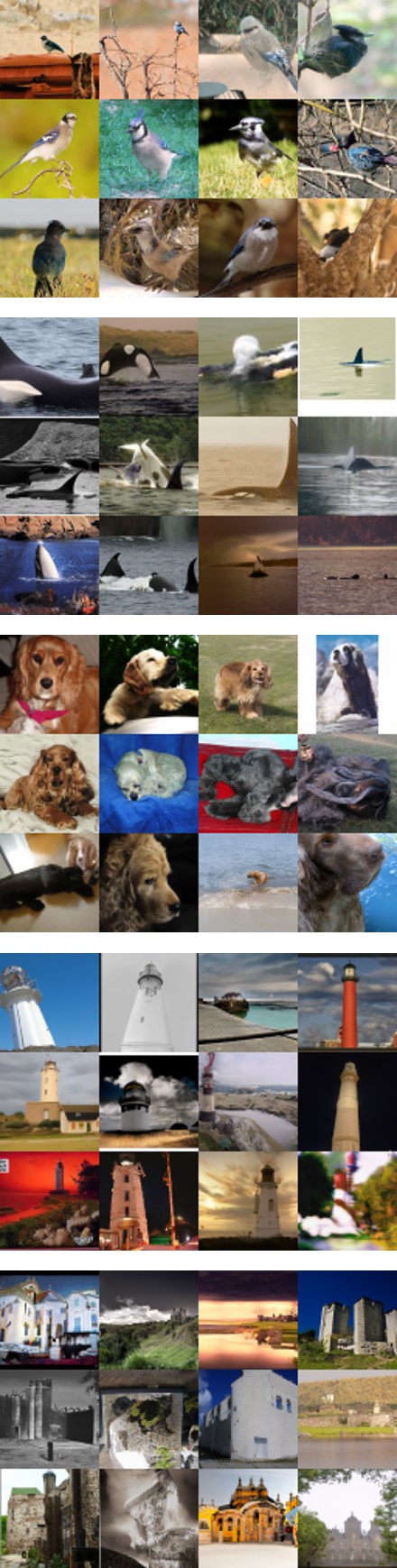}
    \caption{Sehwag \etal~\cite{sehwag2022generating}}
    \end{subfigure}
    \hfill
    \begin{subfigure}[h]{0.321\textwidth}
    \centering
    \includegraphics[width=1.0\columnwidth]{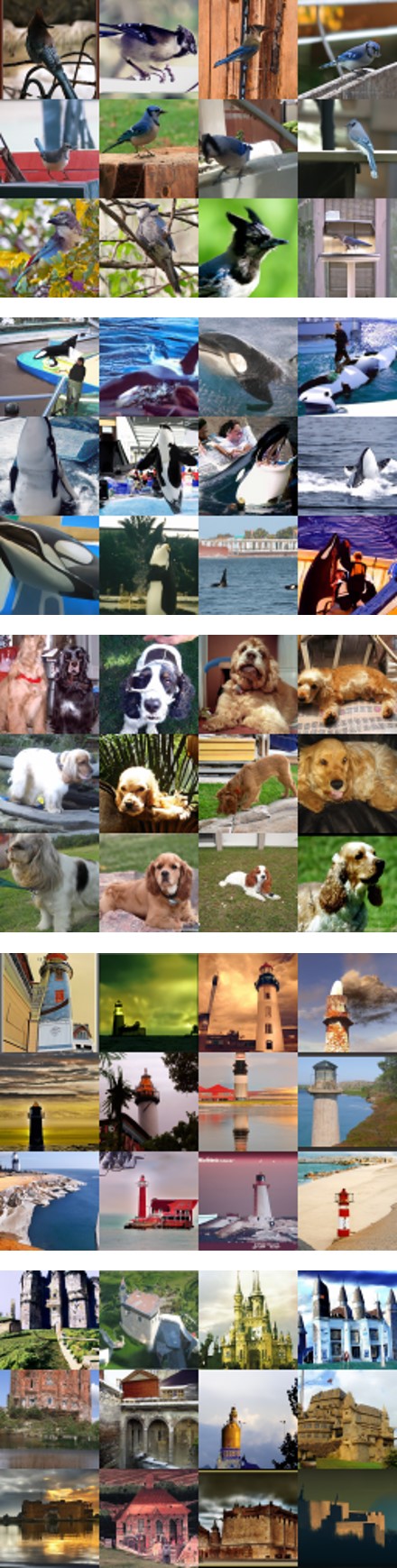}
    \caption{Ours}
    \end{subfigure}
\caption{Sample comparison on ImageNet-64. Generated samples from five classes are exhibited: (i) Jay (top row); (ii) Killer whale (top-middle row); (iii) Cocker spaniel (middle row); (iv) Beacon (middle-bottom row); (v) Castle (bottom row).}
\label{fig:many_imagenet64_samples}
\end{figure}

\begin{figure}[!t]
    \begin{subfigure}[h]{0.321\textwidth}
    \centering
    \includegraphics[width=1.0\columnwidth]{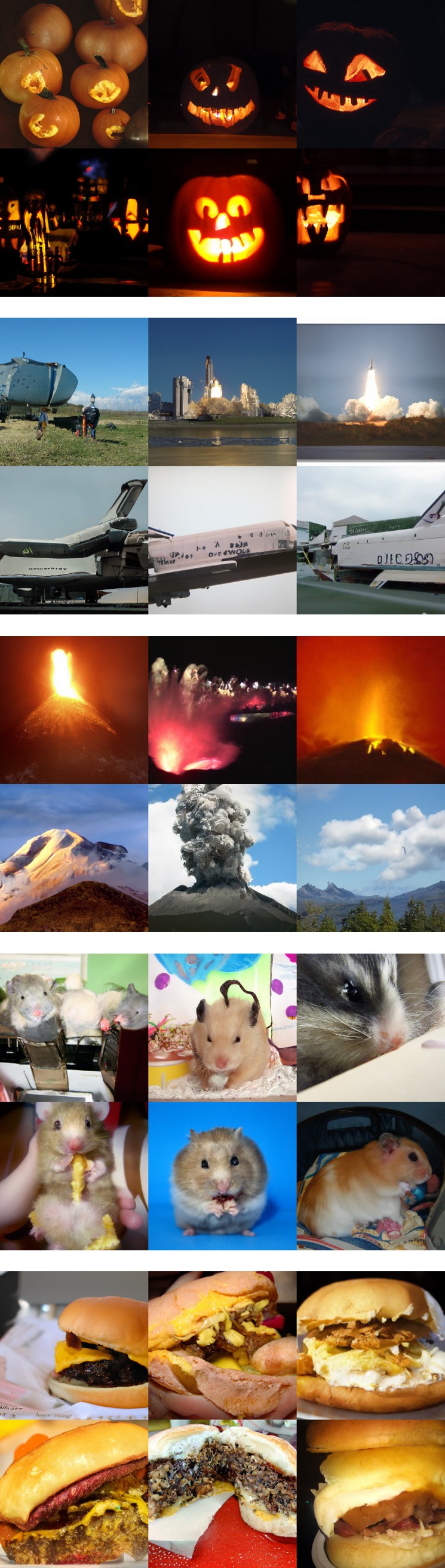}
    \caption{ADM~\cite{dhariwal2021diffusion}}
    \end{subfigure}
    \hfill
    \begin{subfigure}[h]{0.321\textwidth}
    \centering
    \includegraphics[width=1.0\columnwidth]{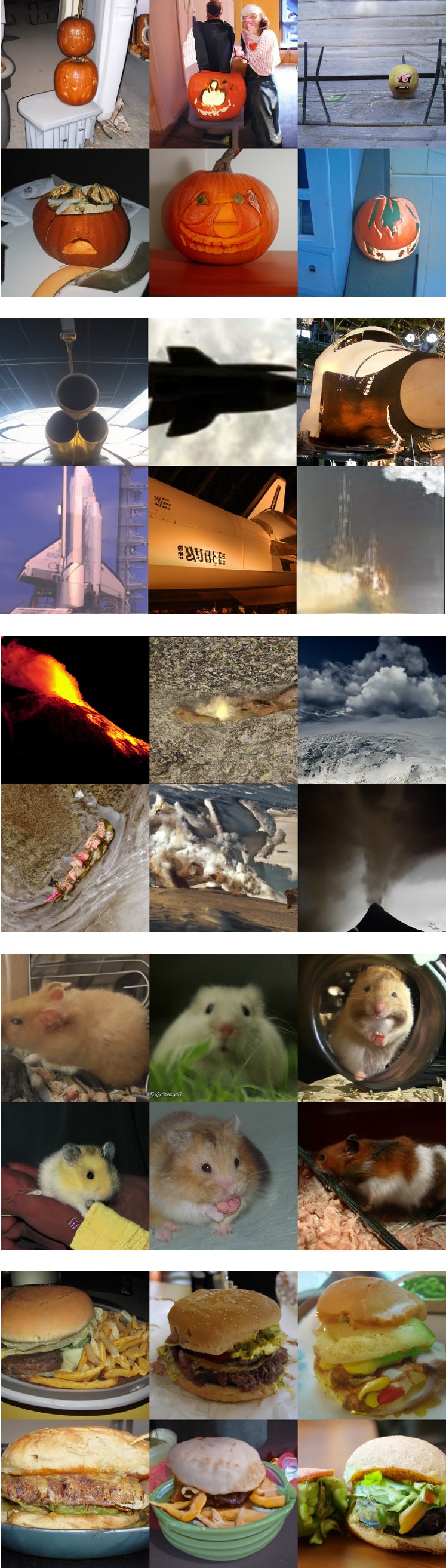}
    \caption{Sehwag \etal~\cite{sehwag2022generating}}
    \end{subfigure}
    \hfill
    \begin{subfigure}[h]{0.321\textwidth}
    \centering
    \includegraphics[width=1.0\columnwidth]{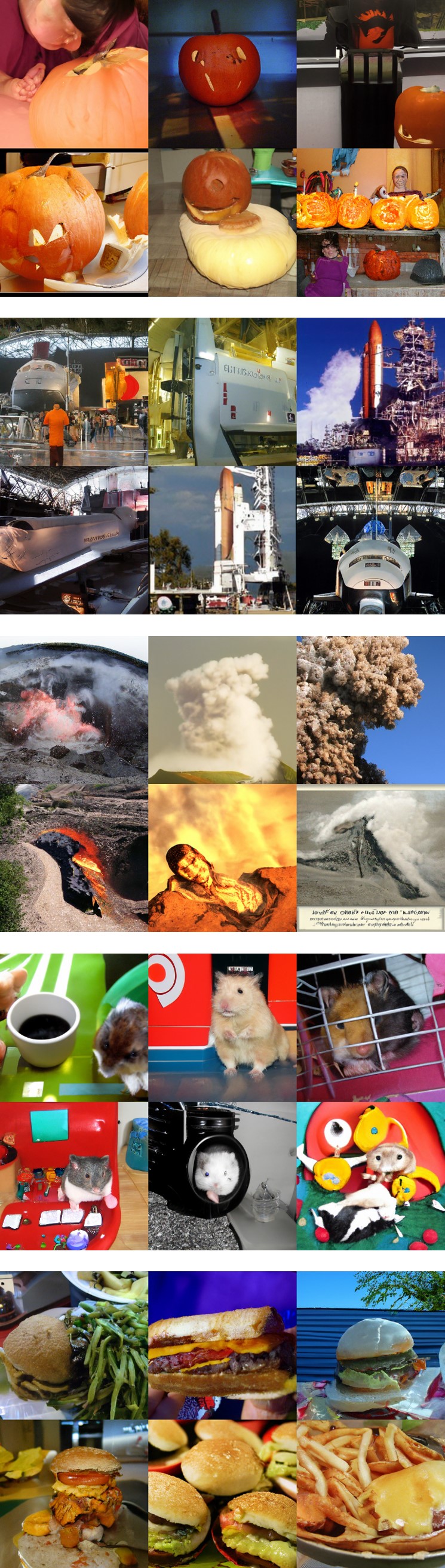}
    \caption{Ours}
    \end{subfigure}
\caption{Additional comparison of generated samples on ImageNet-256. Generated samples from five classes are exhibited: (i) Jack-o`-lantern (top row); (ii) Space shuttle (top-middle row); (iii) Volcano (middle row); (iv) Hamster (middle-bottom row); (v) Cheeseburger (bottom row).}
\label{fig:many_imagenet256_samples}
\end{figure}

\noindent \textbf{Additional generated samples.} To facilitate a more comprehensive qualitative comparison among the samplers, we provide an extensive showcase of generated samples for all the considered datasets. See Figures \ref{fig:many_celeba_samples}--\ref{fig:many_imagenet256_samples} for details.

\newpage

\end{document}